\documentclass[ustc,normalcite]{meta}

\usepackage{xspace}
\usepackage{url}
\usepackage{graphicx}
\renewcommand{\paperlogomark}{%
  \includegraphics[width=\paperlogowidth]{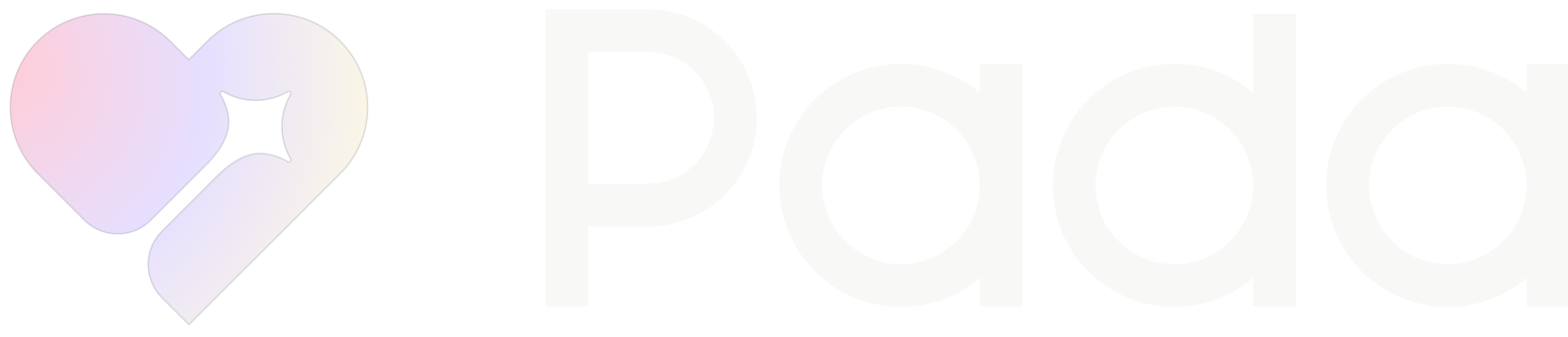}}
\usepackage{amsmath}
\usepackage{amssymb}
\usepackage{tabularx,booktabs,array,colortbl}
\usepackage{float}
\usepackage{algorithm}
\usepackage{algpseudocode}
\usepackage{tcolorbox}
\tcbuselibrary{breakable}

\newtcolorbox{promptbox}{
  breakable,
  colback=brandsoft,
  colframe=brandline,
  boxrule=0.6pt,
  arc=3pt,
  left=6pt,
  right=6pt,
  top=4pt,
  bottom=4pt
}

\newtheorem{hypothesis}{Retrieval Hypothesis}

\definecolor{tablecoral}{HTML}{E8B7AD}
\definecolor{tablecyan}{HTML}{93C8DF}
\definecolor{tableblue}{HTML}{DCE7F6}
\definecolor{tableice}{HTML}{E4F1F5}
\definecolor{tableslate}{HTML}{EDF1F5}
\newcommand{\yesmark}{\textcolor{brandblue}{$\checkmark$}}
\newcommand{\nomark}{\textcolor{tablecoral}{$\times$}}
\newcommand{\namark}{\textcolor[rgb]{0.55,0.58,0.62}{--}}
\newcommand{\degradation}[2]{\cellcolor{tablecoral!#1}#2}
\newcommand{\improvement}[2]{\cellcolor{tablecyan!#1}#2}

\title{Keep It \textcolor{brandblue}{InMind}: Benchmarking the Implicit-Association Blind Spot in Agent Memory}
\author[1,*]{Ruizhe Li}
\author[1,*]{Mingxuan Du}
\author[2]{Benfeng Xu}
\author[1,\dagger]{Zhendong Mao}
\affiliation[1]{University of Science and Technology of China}
\affiliation[2]{Metastone Technology}
\contribution[*]{Equal contribution\quad ${}^{\dagger}$Corresponding author}
\abstract{Long-term memory systems store what a user says in an external store and retrieve it when a related query arrives. This interface rests on an assumption so natural that it is rarely stated: a memory that is needed will resemble the query that needs it. World knowledge breaks the assumption. A tree-nut allergy should change the answer to a macaron request through their almond-flour ingredient, yet the two texts share no cue a retriever can see. We call this failure mode the \emph{implicit-association blind spot} and introduce InMind, a 125-task, expert-verified benchmark spanning ten life domains, with 113 tasks grounded in citable public sources. Its paired controls separate three explanations that existing evaluations conflate: the fact was never stored, the model lacks the bridging knowledge, or the fact was stored and never surfaced. The verdict is clean. With the decisive memory placed in context, the backbone answers 84.0\% of indirect queries; when the same memory must be retrieved, six vector, graph, and agentic memory systems reach at most 14.4\%, even though they recall the same facts on demand at up to 100\%. An embedding with eight times the dimensionality raises answer-blind target recall for every system yet leaves the gap essentially intact. A minimal diagnostic probe that keeps memory visible before the query arrives recovers most of the gap, locating the failure in the query-conditioned interface itself and pointing to \emph{routing}---deciding which facts must stay visible---as the open problem InMind is built to score.}
\metadata[Project]{\href{https://keep-it-inmind.github.io/}{Project Website}\quad\href{https://github.com/imlrz/InMind}{GitHub Repository}}

\begin{document}

\maketitle

\section{Introduction}

\begin{figure}[H]
  \centering
  \includegraphics[width=0.92\textwidth]{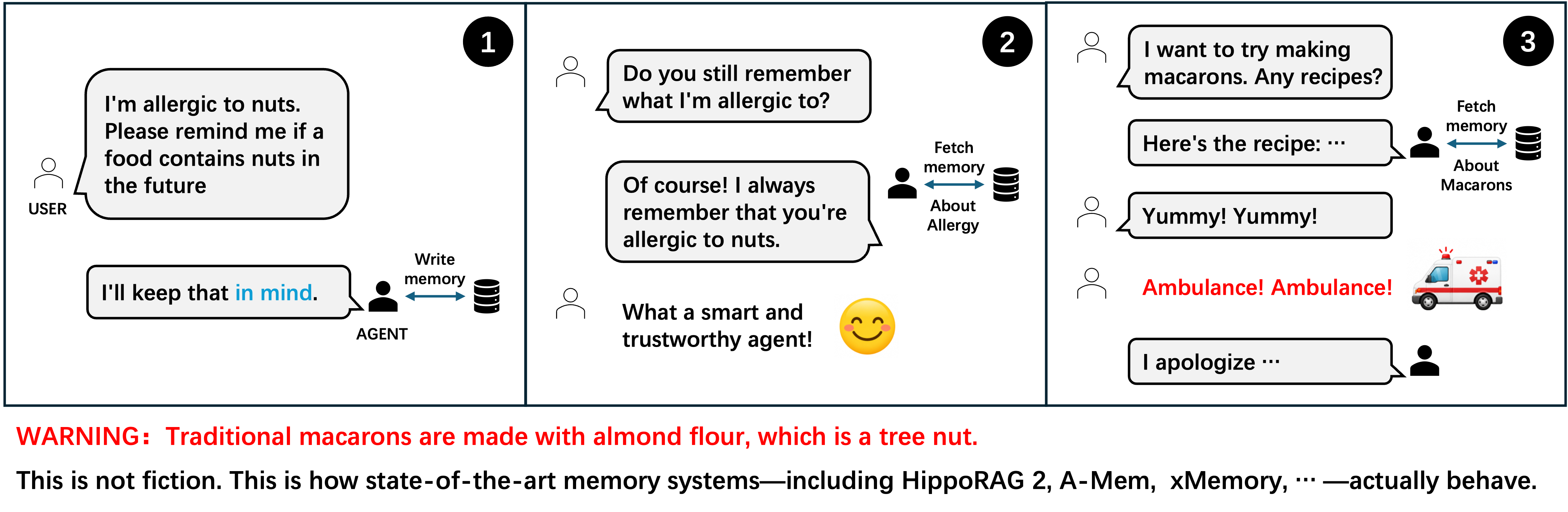}
  \caption{Direct recall does not guarantee memory use: the system recalls the allergy when asked, then fails to apply it to a macaron request. Standard macarons use almond flour, and almond is a major tree-nut allergen \citep{kingArthurMacarons,fdaFoodAllergies}. Appendix~\ref{app:response-examples} records this failure in an evaluated memory system.}
  \label{fig:comic}
\end{figure}

Ask a memory-augmented agent what its user is allergic to, and it answers correctly: tree nuts. Ask it a moment later for a macaron recipe, and it responds with enthusiasm and almond flour. Figure~\ref{fig:comic} illustrates the exchange; Appendix~\ref{app:response-examples} records an evaluated memory system reproducing it verbatim. Nothing was forgotten---the system recalled the allergy on demand seconds earlier. What failed is subtler and, we will argue, structural: the memory was never brought to bear at the one moment it mattered.

Language agents are increasingly expected to act as persistent assistants rather than stateless chatbots \citep{zhang2024memorymechanism,packer2023memgpt,liu2025secom}: a user states a fact about themselves once and expects it to keep mattering days later, in another conversation, on a topic they never connected to it themselves. The dominant way to meet this expectation is retrieval-based memory. A writer distills what the user said into records; when a new query arrives, the system embeds it, retrieves the nearest records, and pastes them into the prompt \citep{lewis2020rag,karpukhin2020dpr,borgeaud2022retro,asai2023selfrag}. The design has real virtues---it scales past the context window, updates modularly, and inherits well-tested machinery from retrieval-augmented generation---and it works on direct questions, because ``what am I allergic to?'' is itself a retrieval cue.

It fails when relevance is created by knowledge that lives in neither text: the allergy note and the macaron request share no token, topic, or embedding-space neighborhood, and what connects them is a fact about French patisserie stated in neither turn. Underneath every retrieval-based design sits what we call the \emph{retrieval hypothesis}, formalized in Section~\ref{sec:retrieval-based}: any memory needed to answer a query lies close to that query under some computable relevance score. Explicit recall satisfies the hypothesis. Implicit associations---drug--food interactions, cross-allergies, dietary law, occupational restrictions, household hazards---violate it, because retrieval commits to a relevance judgment before the model ever sees the memory, and the model is the only component of the system that could recognize the connection. Figure~\ref{fig:comic} is not a hypothetical: HippoRAG~2~\citep{gutierrez2025hipporag2}, A-Mem~\citep{xu2025amem}, xMemory~\citep{wang2025xmemory}, and A-RAG~\citep{du2026arag} all fail exactly this task in the benchmark we introduce below.

The blind spot is doubly hidden. Users calibrate trust by asking the agent what it remembers, and that is precisely the interaction retrieval handles well: an agent that answers direct questions about a fact is indistinguishable, from the outside, from an agent that will use it. Evaluations inherit the same problem in a subtler form: when a system fails an indirect query, three explanations compete---the fact was never stored, the model lacks the knowledge that bridges fact and query, or the fact was stored and never surfaced---and they call for entirely different fixes. Existing benchmarks cannot tell them apart.

We introduce \textbf{InMind}, a 125-task benchmark built to separate them. Every task pairs a personal fact with an indirect query whose correct answer changes because of that fact; 113 bridges are grounded in citable public sources and the remaining 12 are expert-authored, with all 125 expert-verified. Two paired controls and one answer-blind measurement isolate the three explanations: a direct \emph{naive query} tests storage, an \emph{in-context control} tests the backbone's bridging knowledge, and \emph{target recall} tests whether retrieval delivered the fact to the model's context. The verdict is unambiguous. Six memory systems spanning vector, graph, and agentic designs recall the injected facts on demand at up to 100\% and hold the knowledge to bridge them (the in-context backbone scores 84.0\%), yet apply them under indirect queries at no more than 14.4\%, with no query-time configuration exceeding 16.0\%. A final experiment shows what it takes to close the gap: a deliberately minimal \emph{always-in-state} probe, which keeps memory visible before the query arrives instead of retrieving after it, recovers most of it (68.8\%).

Our contributions are as follows:
\begin{itemize}
\item We define the \emph{implicit-association blind spot} and make explicit the retrieval hypothesis that current memory architectures assume and never test.
\item We introduce InMind, a 125-task expert-verified benchmark with 113 source-grounded tasks, whose paired controls separate three failure modes that prior evaluations conflate: storage failure, missing model knowledge, and retrieval failure.
\item We evaluate six memory systems spanning vector, graph, and agentic designs. All recall the injected facts on demand and none applies them reliably under indirect queries; an embedding with eight times the dimensionality raises answer-blind target recall for every system without closing the gap, placing the failure in the retrieval interface itself.
\item We show what it takes to close the gap: a deliberately minimal always-in-state probe---nothing more than a diligently maintained profile---recovers most of it, while hybrid systems that already keep profile-style state alongside retrieval do not. We propose no system; the contrast isolates \emph{routing}, whether decision-critical facts actually reach and survive in the visible state, as the open problem.
\end{itemize}

\section{The Blind Spot in Query-Conditioned Memory}

\subsection{Retrieval-Based Memory and Its Hidden Hypothesis}
\label{sec:retrieval-based}

Let $\mathcal{M}=\{m_1,\ldots,m_n\}$ be a user's stored memories and $q$ a new query. A retrieval-based system selects a subset through a query-conditioned procedure and hands it to the model:
\begin{equation}
\hat{\mathcal{M}} = \mathrm{Retrieve}(\mathcal{M}; q,\theta), \qquad a = \mathrm{LLM}(q,\hat{\mathcal{M}}),
\end{equation}
where $\theta$ collects retriever parameters, indices, scoring rules, or traversal policies. The model reasons only over $\hat{\mathcal{M}}$, so the ordering is decisive: $\mathrm{Retrieve}$ must commit to what is relevant before $\mathrm{LLM}$---the one component holding world knowledge---is ever consulted. Every such system stakes its correctness on an assumption that is rarely stated and, to our knowledge, never tested directly.

\begin{hypothesis}
\label{hyp:retrieval}
If a memory $m \in \mathcal{M}$ is necessary to answer a query $q$, then $m$ is recoverable by a relevance score computed from $q$ alone: $m \in \mathrm{Retrieve}(\mathcal{M}; q,\theta)$ for a suitable choice of $\theta$.
\end{hypothesis}

The quantifier over $\theta$ deserves a caveat. The hypothesis is trivially satisfiable by letting $\mathrm{Retrieve}$ run the full model over every stored memory, but that forfeits the sublinear cost that motivates retrieval and merely relocates the world model into the scoring function. We therefore read the hypothesis as deployed systems do, with $\theta$ ranging over efficient similarity computations. So read, it is unobjectionable for explicit recall, where the query names the memory or a paraphrase of it; we will argue that it is false in general, and false precisely where being wrong is expensive. Section~\ref{sec:embeddings} tests the practical middle ground---embeddings that have absorbed some world knowledge during training---and measures how far it reaches.

\subsection{Implicit Associations}
\label{sec:implicit-association}

We call a memory--query pair $(m,q)$ an \emph{implicit association} when it satisfies three conditions:
\begin{enumerate}
\item \textbf{Necessity}: the memory $m$ is required to answer $q$ safely, correctly, or appropriately.
\item \textbf{Semantic distance}: $m$ and $q$ sit far apart under common lexical, embedding, or topical relevance functions.
\item \textbf{Knowledge bridge}: there exists background knowledge $k$ such that $m \xrightarrow{k} q$.
\end{enumerate}

Consider a user who mentions owning a cat and, weeks later, asks whether lilies would make a good centerpiece. The pair satisfies all three conditions: the memory changes the answer, because many lilies are toxic to cats \citep{aspcaLilies}; the query mentions no animals, pets, or toxicity; and the bridge is veterinary knowledge stated in neither turn. The structure recurs across domains---carbamazepine and grapefruit via a labeled drug interaction \citep{fdaTegretol}, hearing impairment and restaurant choice via noise exposure \citep{oshaNoise}, and Halal dietary rules and medicinal capsules via gelatin origin \citep{wikipediaGelatin}.

\subsection{Why Query-Conditioned Retrieval Fails, and How to Tell}
\label{sec:query-retrieval-fails}

\begin{figure}[t]
  \centering
  \includegraphics[width=0.92\textwidth]{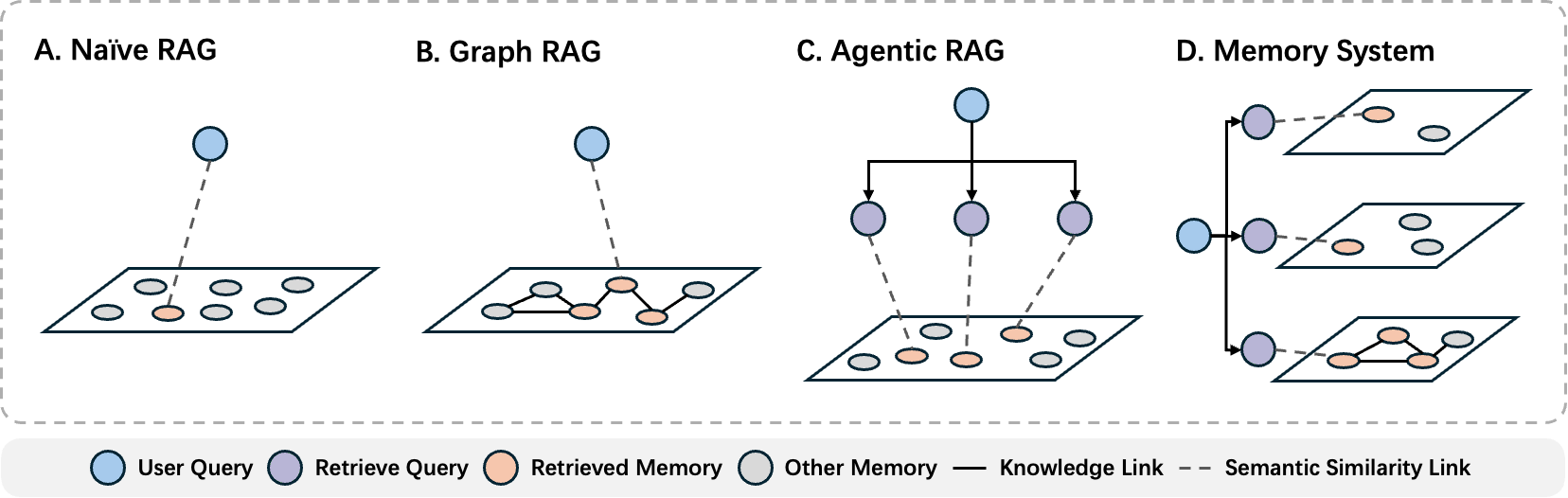}
  \caption{Representative query-time memory architectures. Blue, lavender, and orange denote the user query, retrieval query, and retrieved memory. Every route to the decisive memory crosses at least one semantic-similarity (dashed) link; the solid machinery each design adds sits upstream of that hop and cannot substitute for it.}
  \label{fig:agent-memory-taxonomy}
\end{figure}

Query-conditioned retrieval hands the relevance judgment to a component that lacks the context for making it. A dense retriever places ``cat'' far from ``lily'' because ordinary text treats them as different topics; a sparse retriever finds no shared token; a graph retriever helps only when the bridge already sits in the graph \emph{and} the query activates it. The one component that could apply the bridge---the model---sees memory and query together only after retrieval has already chosen. However memory is organized (Figure~\ref{fig:agent-memory-taxonomy}), this bottleneck is shared: whatever machinery a design adds---knowledge edges in a graph, planned sub-queries in an agentic searcher, hierarchical stores in a memory system---every route to the decisive memory still crosses a semantic-similarity link computed from a representation built before that memory comes into view, and an implicit association is precisely a pair that gives this link no signal.

This account makes a falsifiable prediction with a distinctive shape: (i) with the memory placed directly in context, the model should answer indirect queries well, because the bridging knowledge is there; (ii) under retrieval, explicit recall should stay near ceiling, because storage works; and (iii) indirect application should collapse, because the interface fails. No account based on task difficulty, forgetting, or model weakness predicts this pattern---hard tasks would also defeat the in-context model, forgetting would also defeat direct recall, and a weak model would fail everywhere. InMind is built to test for exactly this signature.

\section{The InMind Benchmark}
\label{sec:bench}

Separating the three explanations imposes requirements that no existing benchmark meets (Section~\ref{sec:related}): the target memory must offer the retriever no similarity cue, or the task tests search quality rather than implicit association; the bridge must rest on verifiable knowledge, or a failure could reflect a contested judgment call rather than a missed connection; and every task needs paired controls, or a wrong answer stays ambiguous among the three explanations. InMind's construction follows from these requirements.

\subsection{Domain Distribution and Sources}

\begin{figure*}[t]
  \centering
  \includegraphics[width=\textwidth]{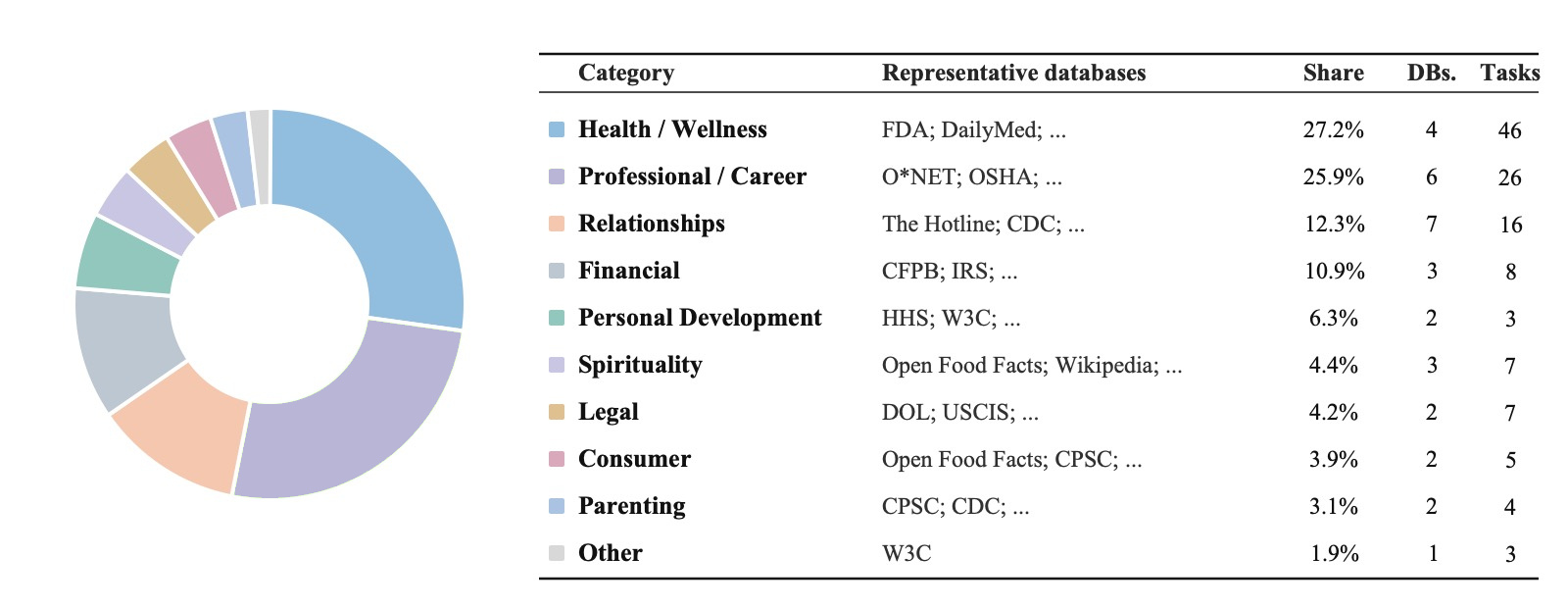}
  \caption{Topic distribution used for sampling (left), and representative databases and final task counts by domain (right).}
  \label{fig:anthropic-distribution}
\end{figure*}

Our domain weights follow Anthropic's analysis of 37,657 personal-guidance conversations \citep{anthropic2026guidance}, retaining even the small \emph{consumer} and \emph{other} slices, which concentrates the benchmark where persistent assistants actually give consequential advice (Figure~\ref{fig:anthropic-distribution}). Within each domain we assemble public, inspectable collections whose content can establish a concrete bridge from a user fact to a later decision---FDA allergen guidance, OSHA regulations, USCIS travel rules, CPSC recalls, and comparable bodies---then sample 3,380 chunks proportionally. Every chunk keeps its source trace (URL and text span), so each source-grounded bridge in InMind is auditable against its public source. Appendix~\ref{app:source-collections} lists all collections.

\nocite{fdaFoodAllergies,dailyMed,cdcPregnancyFoodSafety,medlinePlusHealthTopics,onetDatabase,faaMedications,oshaFacialHair,oshaHeat,oshaNoise,oshaRespiratoryProtection,hotlineIdentifyAbuse,hotlineInternetSafety,hotlineTechAbuse,samhsaRecovery,cdcIPV,loveisrespectDigital,loveisrespectPasswords,cfpbAsk,irsEarlyDistribution,ftcGiftCardScams,hhsPhysicalActivity,w3cUseOfColor,openFoodFacts,wikipediaGelatin,wikipediaIslamicDietaryLaws,dolFmla,uscisTravelDocuments,cpscRecalls,cpscSmallParts,cdcChokingHazards}

\subsection{Task Extraction}

Every source-grounded task begins from a source chunk. An extractor reads one chunk and first judges whether it can support a valid task; extractable chunks become a record holding (i) a user message stating a personal fact, (ii) a direct \emph{naive query} asking for that fact, (iii) an indirect \emph{query} whose answer should change because of it, and (iv) a source-grounded explanation of the bridge. The prompt bars the user message from disclosing the later object, bars the indirect query from naming the fact or an obvious synonym, and requires the chunk itself to support the bridge (Appendix~\ref{app:generation-prompt}). Gemini~2.5 Flash \citep{google2025gemini25} serves as the main extractor, with Claude Sonnet~4.6 \citep{anthropic2026sonnet46}, Claude Opus~4.8 \citep{anthropic2026opus48}, and DeepSeek~V4 Pro \citep{deepseek2026v4} broadening generation; together they produce 1,000 candidates.

Pairing each fact with both a naive and an indirect query is what separates the two abilities the literature conflates: a system that answers the naive query and fails the indirect one demonstrably holds the fact---it failed to surface it when it counted.

\subsection{Filtering and Human Review}
\label{sec:filtering}

Three content filters reduce the 1,000 candidates to the evaluation set; Appendix~\ref{app:filter-details} states each rule in full.

\begin{figure*}[t]
  \centering
  \includegraphics[width=\textwidth]{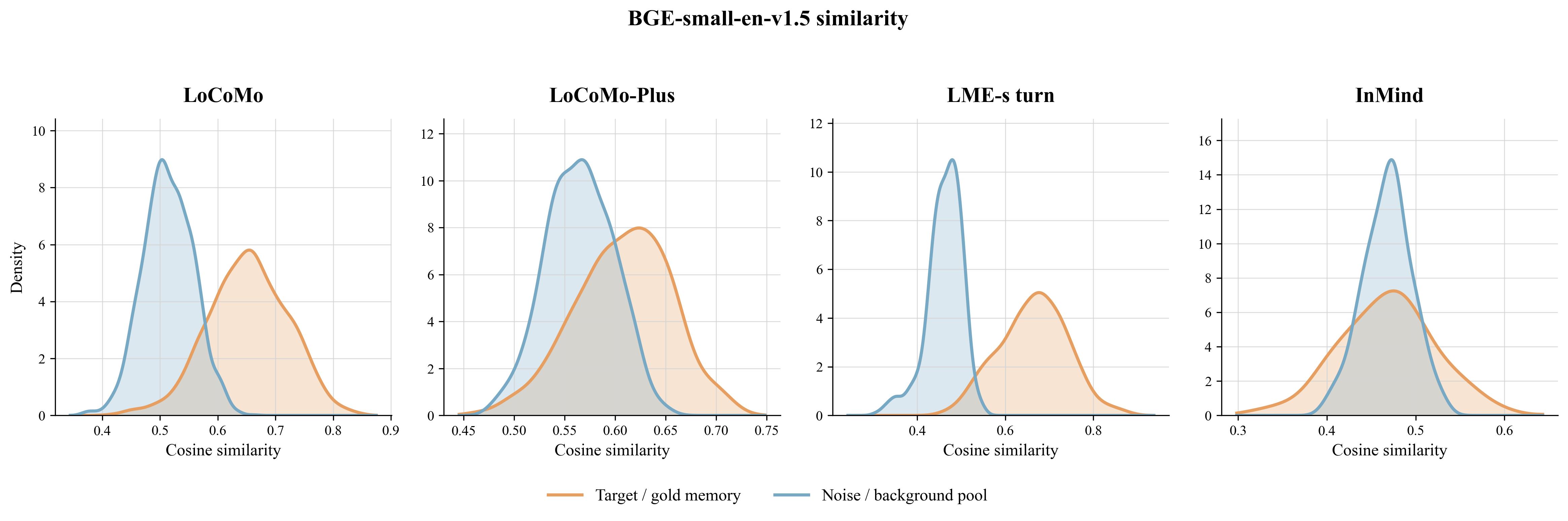}
  \caption{Query--memory similarity under BGE-small-en-v1.5, an encoder not used for task filtering. Orange denotes target memories and blue the background pool. Only InMind shows virtually no target--noise separation.}
  \label{fig:similarity}
\end{figure*}

\textbf{(i) Similarity.} BM25 \citep{robertson2009bm25} and MiniLM \citep{wang2020minilm,reimers2019sentencebert} score every memory--query pair, and we discard the 700 candidates whose target reads as an obvious lexical or dense match to its query, since an ordinary retrieval cue disqualifies a pair as an implicit association. To avoid evaluating this filter with its own encoder, we re-embed InMind and three comparison benchmarks---LoCoMo \citep{maharana2024locomo}, LoCoMo-Plus \citep{li2026locomoplus}, and LME-s from LongMemEval \citep{wu2024longmemeval}---with BGE-small-en-v1.5~\citep{xiao2023cpack}, an encoder not used in filtering (Figure~\ref{fig:similarity}). InMind alone places the decisive memory no closer to the query than its distractors, leaving a similarity retriever no signal to exploit.

\textbf{(ii) Conflict.} All tasks share one background conversation trace (Section~\ref{sec:eval-setting}), so an LLM discards any injected fact that contradicts the persona the trace already establishes---a cat owner in a history where the user has no pets---which would make a task incoherent instead of difficult.

\textbf{(iii) Expert verification.} Human experts confirm that the bridge is factually correct, that the fact materially changes the answer, and that no overt retrieval cue survives; they may retain, revise, or reject each candidate.

These filters share a property worth stating plainly: \emph{each depends only on task content and the fixed background trace, and none consults any system's output}. Retention therefore cannot be influenced by whether a system answered correctly, and scoring the retained set is equivalent to having run only that set. The conflict and expert filters reduce the 300 surviving candidates to 113; together with a small hand-written expert component covering domains where public sources run thin, this yields the \textbf{125-task evaluation set}: 113 source-grounded and 12 expert-authored tasks, so all but twelve trace to a citable public source carrying an expert-verified bridge.

Many tasks are safety-relevant by design: a missed tree-nut allergy is unambiguously wrong where a missed preference stays arguable, and we chose the region where failure can be scored without dispute. The same associative structure appears in personalization, etiquette, compliance, and planning (Appendix~\ref{app:domain-examples}).

\subsection{Evaluation Protocol}
\label{sec:eval-setting}

A memory benchmark must make the memory survive realistic interference before it is needed. For each task, we insert its generated memory turn into the middle of the fixed 47-session LongMemEval-s (LME-s) conversation trace \citep{wu2024longmemeval} and continue the simulated dialogue so that each memory system operates normally: the injected fact must survive 38 further sessions of ordinary interaction before it is ever queried. After this conversation, we test the system twice---with the task's direct \emph{naive query}, then with its indirect \emph{query}. The complete construction and injection schedule are given in Appendix~\ref{app:injection-protocol}. All scores are computed on the 125 tasks retained by the content filters of Section~\ref{sec:filtering}, which are independent of every system's output.

We report two complementary measurements on the indirect query. \emph{Target recall} is an answer-blind, binary post-hoc judgment of whether the answerer's actual context contains the target personal fact; the judge receives the memory, context, query, and bridge explanation, but never the generated answer. \emph{Indirect application} is a binary, context-aware answer judgment that assesses whether the response follows the source-grounded bridge (Appendix~\ref{app:judge-prompts}). Separating them identifies whether an error arises before the model can see the fact or after it has seen it.

\section{Experiments: Locating the Failure}
\label{sec:experiments}

\subsection{Systems}

\textbf{In-context backbone control.} Places the target memory and the query directly in context, measuring whether the model holds the world knowledge to bridge them once both are visible.

\textbf{Retrieval-based memory.} Six systems---A-RAG, xMemory, Mem0, A-Mem, HippoRAG~2, MemoryOS---each run with MiniLM (384-dim) and text-embedding-3-large (emb3-large, 3,072-dim) \citep{wang2020minilm,reimers2019sentencebert,openai2024embeddings}. We add a single-shot Naive RAG control over raw LME-s turn chunks under MiniLM, emb3-large, or BM25 \citep{robertson2009bm25}. All share the 125-task set, with GPT-5-mini \citep{openai2025gpt5} as answerer and binary judge; hyperparameters appear in Appendix~\ref{app:system-hparams}.

\subsection{Main Results}

Table~\ref{tab:main} states the result in one line: naive recall reaches up to 100.0\%, indirect application never exceeds 16.0\%. The naive column runs high almost everywhere, so the memory-write and direct-retrieval pipelines work and nothing here is forgetting. The application column inverts that picture. Every retrieval configuration---vector stores, graph traversal, an agentic searcher with a fifteen-step budget---lands between 3.2\% and 16.0\%, and the six memory systems themselves top out at 14.4\%, below even the Naive RAG control at 16.0\%. The spread among them is smaller than the distance separating all of them from the 84.0\% backbone, which makes the blind spot a property of the paradigm rather than of any one implementation. The remainder of this section works through the three candidate explanations in turn.

\begin{table*}[t]
\centering
\caption{Results (\%) on the 125-task InMind benchmark: direct \emph{Naive} recall, answer-blind \emph{Target Recall} in the indirect-query context, and end-to-end \emph{Application}. The $\Delta$ columns report percentage-point differences (Recall--Naive, Application--Recall, Application--Naive); muted coral and cyan-blue indicate negative and positive changes, with darker shading for larger magnitudes. The backbone receives the target memory in context and therefore has no naive-retrieval score.}
\label{tab:main}
\scriptsize
\setlength{\tabcolsep}{1pt}
\begin{tabular}{lcccccc}
\toprule
System & Naive & Target Recall & Application & $\Delta_{\text{Recall--Naive}}$ & $\Delta_{\text{App.--Recall}}$ & $\Delta_{\text{App.--Naive}}$ \\
\midrule
\rowcolor{tableblue}
\multicolumn{7}{c}{\emph{Backbone Control}} \\
Backbone (GPT-5-mini) & -- & \textbf{100.0} & \textbf{84.0} & -- & \degradation{15}{-16.0} & -- \\
\midrule
\rowcolor{tableslate}
\multicolumn{7}{c}{\emph{Naive RAG}} \\
Dense (MiniLM)~\citep{wang2020minilm} & 92.0 & 0.8 & 9.6 & \degradation{85}{-91.2} & \improvement{44}{+8.8} & \degradation{77}{-82.4} \\
Dense (emb3-large)~\citep{openai2024embeddings} & \textbf{97.6} & 6.4 & \textbf{16.0} & \degradation{85}{-91.2} & \improvement{48}{+9.6} & \degradation{76}{-81.6} \\
BM25~\citep{robertson2009bm25} & 53.6 & 3.2 & 9.6 & \degradation{47}{\textbf{-50.4}} & \improvement{32}{+6.4} & \degradation{41}{\textbf{-44.0}} \\
\midrule
\rowcolor{brandbg}
\multicolumn{7}{c}{\emph{MiniLM}} \\
A-RAG~\citep{du2026arag} & 97.6 & 5.6 & 4.8 & \degradation{86}{-92.0} & \degradation{1}{-0.8} & \degradation{87}{-92.8} \\
xMemory~\citep{wang2025xmemory} & 84.8 & 3.2 & 4.8 & \degradation{76}{-81.6} & \improvement{8}{+1.6} & \degradation{75}{-80.0} \\
Mem0~\citep{chadha2025mem0} & 76.0 & 2.4 & 3.2 & \degradation{69}{-73.6} & \improvement{4}{+0.8} & \degradation{69}{-72.8} \\
A-Mem~\citep{xu2025amem} & \textbf{99.2} & 2.4 & 6.4 & \degradation{90}{-96.8} & \improvement{20}{+4.0} & \degradation{87}{-92.8} \\
HippoRAG~2~\citep{gutierrez2025hipporag2} & 93.6 & 0.8 & 8.8 & \degradation{87}{-92.8} & \improvement{40}{+8.0} & \degradation{79}{-84.8} \\
MemoryOS~\citep{kang2025memoryos} & 87.2 & 2.4 & 8.0 & \degradation{79}{-84.8} & \improvement{28}{+5.6} & \degradation{74}{-79.2} \\
\midrule
\rowcolor{tableice}
\multicolumn{7}{c}{\emph{text-embedding-3-large}} \\
A-RAG~\citep{du2026arag} & 93.6 & 11.2 & 7.2 & \degradation{77}{-82.4} & \degradation{4}{-4.0} & \degradation{81}{-86.4} \\
xMemory~\citep{wang2025xmemory} & 91.2 & 6.4 & 6.4 & \degradation{79}{-84.8} & 0.0 & \degradation{79}{-84.8} \\
Mem0~\citep{chadha2025mem0} & 76.8 & 6.4 & 6.4 & \degradation{66}{-70.4} & 0.0 & \degradation{66}{-70.4} \\
A-Mem~\citep{xu2025amem} & \textbf{100.0} & \textbf{12.0} & 9.6 & \degradation{82}{-88.0} & \degradation{2}{-2.4} & \degradation{85}{-90.4} \\
HippoRAG~2~\citep{gutierrez2025hipporag2} & 96.0 & 2.4 & 5.6 & \degradation{87}{-93.6} & \improvement{16}{+3.2} & \degradation{85}{-90.4} \\
MemoryOS~\citep{kang2025memoryos} & 96.8 & 7.2 & 14.4 & \degradation{84}{-89.6} & \improvement{36}{+7.2} & \degradation{77}{-82.4} \\
\bottomrule
\end{tabular}
\end{table*}

\subsection{Not the Model: In Context, the Backbone Applies the Bridge}
\label{sec:finding-backbone}

The first candidate explanation is that the tasks are simply hard---that no model could connect a months-old fact to an unrelated-looking question. The in-context control rules this out. With the fact visible, GPT-5-mini reaches 84.0\% on indirect queries (105/125): the same model, on the same tasks, succeeds five times out of six, and the 95\% Wilson interval around the backbone, $[76.6, 89.4]$, sits far above the interval around the best retrieval configuration, $[10.6, 23.4]$. What separates 84.0\% from 16.0\% is access, not ability. The difficulty sits strictly upstream, where something must decide to surface the cat memory before the model can reason at all---and that decision falls to a component incapable of reasoning. The world knowledge that would mark the memory as relevant is locked out of the moment relevance gets decided.

\subsection{Not Storage: Retrievable on Demand, Absent When It Matters}
\label{sec:finding-storage}

The second candidate explanation is that the fact was never stored or did not survive the intervening sessions. The naive column refutes it: most systems answer a direct request for the fact almost perfectly, A-Mem reaching 100.0\%, so the fact was written, survived 38 sessions of interference, and stays retrievable on demand. The target-recall column then locates the failure precisely. Under an indirect query, the decisive fact reaches the answerer's context in only 0.8--5.6\% of MiniLM runs and 2.4--12.0\% of emb3-large runs. The fact is stored, the model can use it, and it is absent from the model's context at the only moment it matters. Appendix~\ref{app:response-examples} shows this verbatim for Task~155, where xMemory recalls the tree-nut allergy on demand and then recommends macarons.

The consequence inverts the usual relation between a score and trust: the interaction users test is the one that works, while the interaction that depends on the memory runs near chance. An agent that recalls an allergy on request and recommends macarons unprompted has earned trust it does not merit.

\subsection{Not Representation: Better Embeddings Help but Cannot Close the Gap}
\label{sec:embeddings}

The last candidate explanation within the paradigm is weak representation: perhaps a stronger embedding would place ``grapefruit'' near the drugs it interacts with, and the blind spot would dissolve with scale. Swapping MiniLM (384-dim) for text-embedding-3-large (3,072-dim) tests this. Retrieval does improve for every system: target recall rises across all six (A-Mem 2.4\%$\to$12.0\%, A-RAG 5.6\%$\to$11.2\%, MemoryOS 2.4\%$\to$7.2\%, and the rest in step). End-to-end application follows less cleanly: five of six improve---MemoryOS 8.0\%$\to$14.4\%, A-Mem 6.4\%$\to$9.6\%, A-RAG 4.8\%$\to$7.2\%, xMemory 4.8\%$\to$6.4\%, Mem0 3.2\%$\to$6.4\%---while HippoRAG~2 declines (8.8\%$\to$5.6\%), and each individual shift amounts to a few tasks out of 125, inside the $\pm$4--5 point binomial confidence interval at these accuracies (A-RAG's two indices are additionally built separately; Appendix~\ref{app:system-hparams}). We therefore read the aggregate direction as real and the per-system ordering as noise.

A plausible mechanism explains both the gain and the ceiling. A stronger embedding has itself absorbed world knowledge during training, so its similarity function already encodes some of the bridges we ask about, and grapefruit drifts a little closer to the drugs it interacts with. But an embedding bridges an association only where the relation left a distributional trace in training text, and the bridges that decide these tasks are pharmacological, legal, religious, or developmental---most were never written down as co-occurrence. An eightfold capacity increase recovers a few points of a seventy-point gap: absorbing world knowledge into a similarity score approximates a world model from the wrong direction, and the residual is what InMind measures.

\section{What It Takes to Close the Gap}
\label{sec:always-in-state}

Section~\ref{sec:experiments} eliminated the model, the storage, and the representation, leaving the query-conditioned interface itself. Two questions follow. Can the gap be closed from within the paradigm, by retaining more and searching harder? And if not, what does close it? We take them in order; the answers bracket an open problem: routing.

\subsection{Why Searching Harder Cannot Substitute}
\label{sec:searching-harder}

Retrieval-based memory is plainly necessary: an external store scales past the context window and preserves a fine-grained record. The question is sufficiency, and the strongest case for it comes from General Agentic Memory (GAM), which keeps the complete history and makes search the core runtime operation, with an agentic researcher that plans, searches, integrates, and reflects over a full page store \citep{yan2025gam}. Lossless retention and runtime search do solve the failure they target, which is forgetting. But our systems have already cleared that bar---their naive accuracy proves the fact is retained---and they still fail. Retention gives a retriever nothing to act on when no signal tells it to look: as Section~\ref{sec:query-retrieval-fails} argues, the query representation is fixed before the decisive fact comes into view, so searching a larger haystack more thoroughly leaves the needle looking nothing like the query. A-RAG makes this concrete: it plans, issues keyword and semantic searches, reads chunks, loops up to fifteen times, and scores 4.8\% and 7.2\%, among the lowest here. Iteration cannot rescue a search whose every query issues from the same blind representation.

Retrieval alone is therefore insufficient, and the ability an agentic searcher must acquire is more specific than asking better questions. A memory store is not an oracle: it returns what resembles the query it receives, so even the bridge-aware ``what about this user could make lilies dangerous?'' retrieves nothing---no stored fact resembles that question either. The probe that succeeds is ``does the user keep a cat?'', phrased in the vocabulary of the stored fact rather than of the request; to issue it, a searcher must first recall, unprompted, that lilies are toxic to cats, then single cats out among everything lilies might endanger---allergic housemates, birds, curious toddlers---and query each candidate in turn. This is what it means to \emph{hypothesize the bridge}, and every hypothesis is a guess made before anything in the store confirms it: the enumeration is probabilistic exactly where safety demands reliability, which is why we treat it as an open direction rather than a fix.

\subsection{Keeping Memory Visible Recovers the Gap}

If the query-conditioned interface causes the failure, then removing it---while changing nothing else about the task---should recover most of the gap, even for an unsophisticated method. The claim is falsifiable, so we test it.

An \emph{always-in-state} memory departs from Section~\ref{sec:retrieval-based} in one respect: a representation $s_t$ reaches the model before the query arrives, and no query-time retriever selects it:
\begin{equation}
a = \mathrm{LLM}(q, s_t).
\end{equation}
The state may be an injected profile, a persistent summary, or a latent memory state; persistent visibility defines it, whatever the format. The model reasons over $s_t$ and $q$ jointly, discovering a cat--lily bridge with no mismatched query standing in the way. We emphasize that this design is neither ours nor rare: profile-style state is a standard component of deployed assistants, and practical memory systems are typically \emph{hybrids} that keep such a state alongside a retrieved store---MemoryOS, evaluated above, already places a user profile and knowledge entries in every answer context \citep{kang2025memoryos}. What has been missing is measurement: no benchmark tells a hybrid whether its persistent slice actually carries the facts that must stay visible.

To isolate the variable, we run the design at its barest: one markdown file, capped at 200 lines, rewritten by a GPT-5-mini updater after each session and prepended whole to the system prompt at answer time---no embeddings, no vector store, no index, no ranking, no query-time retrieval (Appendix~\ref{app:always-in-state}). This is nothing more than a diligently maintained profile, and we propose it for nothing: a 200-line file is no architecture---its state needs rewriting every interaction, and its finite context budget makes facts crowd each other out as volume grows. Table~\ref{tab:always-in-state} reports the result: 68.8\% on indirect queries, against at most 16.0\% for any query-time configuration, with direct recall matched at 98.4\%.

\begin{table}[H]
\centering
\small
\caption{Always-in-state diagnostic on the same 125 tasks as Table~\ref{tab:main}.}
\label{tab:always-in-state}
\begin{tabular}{lccc}
\toprule
Method & Naive & Indirect & $\Delta_{\text{Indirect--Backbone}}$ \\
\midrule
Backbone (GPT-5-mini) & -- & 84.0 & 0.0 \\
Best retrieval configuration & 97.6 & 16.0 & -68.0 \\
Always-in-state (GPT-5-mini) & 98.4 & \textbf{68.8} & \textbf{-15.2} \\
\bottomrule
\end{tabular}
\end{table}

The value here is diagnostic, and we state it carefully. The probe is not a controlled ablation of any one system---it differs in state format and in updater as well---but it demonstrates sufficiency: a design whose memory is visible before the query arrives, with nothing more than competent write-time curation behind it, recovers most of what the entire retrieval paradigm loses. What separates it from every row of Table~\ref{tab:main} is that no query-time retriever stands between the model and the memory, and that one design change outweighs the accumulated apparatus of vector stores, knowledge graphs, and agentic search. Hypothesis~\ref{hyp:retrieval} is what fails; the retriever works as designed.

The comparison with MemoryOS sharpens the point. Carrying an always-visible profile is not by itself enough: MemoryOS posts the strongest application score among the six memory systems (14.4\% with emb3-large) and still sits more than fifty points below the probe, and its answer-blind target recall (7.2\%) shows the decisive fact rarely reached even its persistent slice. Taken together, the probe and MemoryOS pin down the real variable: not persistent visibility as such, but \emph{what is visible when the query arrives}. The question a hybrid must answer is therefore not whether to keep a profile---deployed systems already do---but what earns a place in it, which is the routing problem Section~\ref{sec:routing} takes up.

\subsection{Routing as the Open Problem}
\label{sec:routing}

The two designs are complementary necessities---retrieval preserves a lossless record at scale, always-visible state lets distant knowledge-connected facts be reasoned over---and hybrids combining them already exist. What they lack is a criterion. Every hybrid embeds a routing decision: which memories earn persistent state at write time, and when a query justifies an expensive search over the fine-grained record at read time. Today that decision falls to proxies---recency, frequency, MemoryOS's heat-thresholded promotion---none of which can see the distant bridge that will one day make a fact decision-critical; and the right criterion is vertical-specific besides, since a medical assistant, a financial advisor, and a coding agent price the same forgotten fact very differently. This is the setting InMind is built to serve: to our knowledge it is the first benchmark that scores the routing decision itself, because a hybrid whose router leaves a decision-critical fact out of the visible state fails its indirect query.

Routing defers the problem rather than solving it. Facts too unimportant to earn persistent state can still turn decision-critical later through a distant bridge, and they are exposed the moment the system falls back on ordinary retrieval. Closing that gap needs a relevance function of a different kind (Section~\ref{sec:conclusion}).

\section{Related Work}
\label{sec:related}

\textbf{Memory for language agents.}
Early systems retrieve observations to sustain simulated behavior or accumulate skills \citep{park2023generative,wang2023voyager}, and MemGPT separates a limited context from external storage \citep{packer2023memgpt}. Later work extends the pipeline: vector stores distill compact reusable facts \citep{zhong2024memorybank,chadha2025mem0}; temporal and self-evolving memories revise records over time \citep{wang2024recallm,liu2025secom,wang2024sage,chen2025meminsight}; graph systems expose relational structure for multi-hop retrieval \citep{gutierrez2024hipporag,gutierrez2025hipporag2,strang2025zep}; heterogeneous stores separate memory types before aggregation \citep{xu2025amem,wang2025xmemory}. Across this line of work, the query-then-retrieve interface inherited from retrieval-augmented generation remains the default \citep{lewis2020rag,karpukhin2020dpr,borgeaud2022retro,asai2023selfrag,yan2024crag,jeong2024adaptiverag,sarthi2024raptor,edge2024graphrag}. However these designs organize memory, they share the inference pattern of Figure~\ref{fig:agent-memory-taxonomy}: retrieval fires from a representation built before the model can see the decisive memory. These efforts improve what an agent stores; application is the gap we isolate.

\textbf{Memory benchmarks.}
Long-term-memory benchmarks ask whether an agent uses information from earlier interactions. Standard formulations test explicit recall, where query and target share a direct cue; state aggregation, where temporally ordered memories combine into a current state; or entity-based multi-hop reasoning, where an intermediate memory links query to answer \citep{wu2024longmemeval,maharana2024locomo,hu2025mabench}; Appendix~\ref{app:benchmark-taxonomy} illustrates these settings. Recent work reaches past factual recall. ImplicitMemBench evaluates behavioral implicit memory---procedural learning, priming, conditioning---in a short Learning--Interfere--Test protocol \citep{qin2026implicitmembench}. LoCoMo-Plus evaluates \emph{cognitive} memory: whether latent conversational constraints such as a user's state, goal, or value survive cue--trigger semantic disconnect \citep{li2026locomoplus}. InMind targets knowledge-mediated application of an \emph{explicit} user fact, where relevance rests on an external, verifiable bridge that the conversation never states.

Construction provenance separates it further. Where a benchmark's associations are authored by an LLM---as in the simulated dialogues and generated tasks common in this space \citep{maharana2024locomo,li2026locomoplus}---each association is by definition one models find natural, and the same prior that proposed it at construction time will tend to re-derive it at test time, so such benchmarks preferentially sample the connections models already make unaided. InMind's source-grounded bridges come instead from public documents: a task enters the pool because a citable source asserts the connection, whether or not any model finds it salient, and the extractor is barred from supplying knowledge beyond its source chunk (Appendix~\ref{app:generation-prompt}). The association distribution is thereby anchored to expert knowledge rather than to model priors. Table~\ref{tab:benchmark-comparison} adds the properties that distinguish it: the knowledge bridge, source grounding, and a paired direct-recall control that separates a stored-but-unsurfaced fact from one never stored. In-context upper bounds we leave untabulated because looser analogues are common---LoCoMo and LongMemEval report full-history baselines, ImplicitMemBench an oracle-injection analysis, LoCoMo-Plus full-context backbone runs---though these bundle long-context search together with knowledge application, where InMind's paired control isolates application by presenting the target memory alone. In the table, ``knowledge bridge'' means external, verifiable knowledge absent from the conversation, and ``source-grounded'' that tasks are anchored to citable external documents (113 of InMind's 125 tasks; the remainder are expert-authored, and every bridge is expert-verified either way).

\begin{table}[t]
\centering
\scriptsize
\setlength{\tabcolsep}{3pt}
\renewcommand{\arraystretch}{1.12}
\caption{Comparison of memory benchmarks. \yesmark: supported as a paired, per-task property; \nomark: absent; \namark: not applicable to the benchmark's construct (the main task is itself direct recall, or the design deliberately excludes external knowledge). The control column marks a paired diagnostic supplied by the benchmark, not merely the presence of recall questions.}
\label{tab:benchmark-comparison}
\begin{tabular}{p{0.21\textwidth}p{0.18\textwidth}cccc}
\toprule
Benchmark & Memory type & \shortstack{Long\\context} & \shortstack{Knowledge\\bridge} & \shortstack{Source\\grounded} & \shortstack{Direct-recall\\control} \\
\midrule
LoCoMo / LongMemEval & Factual & \yesmark & \nomark & \nomark & \namark \\
ImplicitMemBench & Behavioral implicit & \nomark & \namark & \nomark & \nomark \\
LoCoMo-Plus & Cognitive & \yesmark & \nomark & \nomark & \nomark \\
\rowcolor{brandbg}\textbf{InMind} & \textbf{Knowledge-mediated} & \yesmark & \yesmark & \yesmark & \yesmark \\
\bottomrule
\end{tabular}
\end{table}

\section{Limitations}

The benchmark is diagnostic, and its 125 tasks lean toward health, wellness, and safety, where failure is easiest to judge, leaving humor, etiquette, long-term goals, and institutional policy uncovered. At $n=125$ a few percentage points are sampling noise; our conclusions rest on sixty-to-seventy-point effects. GPT-5-mini serves as both answerer and judge, a self-preference risk that could inflate absolute scores. The bias applies to the backbone control and the retrieval systems alike and cannot manufacture the gap between them, and the expert audit in Appendix~\ref{app:judge-audit} bounds the error of each judge; still, an independent judge model remains the most significant methodological gap.

We also assume each bridge is factual and uncontested, where real ones can be probabilistic, jurisdiction-dependent, or disputed; and an agent optimized for InMind could over-warn without our rubric seeing it, since applying the bridge earns credit and over-eagerness costs nothing. Measuring that needs negative controls, left to future work.

\section{Conclusion}
\label{sec:conclusion}

A memory can be essential to a query without resembling it, its relevance created by world knowledge the retriever lacks. On InMind this blind spot costs the leading memory systems the difference between 84.0\% and 16.0\%, while those same systems recall those same facts on demand at up to 100\%. The uncomfortable reading is that the field has spent its effort where the problem already worked: storage, indexing, graph construction, and iterative search all improve what an agent finds once it knows what to look for---and that is the whole difficulty. Query-conditioned retrieval asks a similarity function for a judgment that requires a world model, then consults the world model afterwards. Until a relevance function conditioned on knowledge exists, an agent's memory stays reliable where the user could have told it again, and unreliable where they assumed it was kept in mind.

\section*{Ethics Statement}

InMind contains no personal data, and every user fact is synthetic. Of its 125 tasks, 113 are grounded in public sources, including institutional guidance from FDA, OSHA, CPSC, USCIS, and CDC as well as public reference databases; the remaining 12 are expert-authored, and every bridge is expert-verified. Some tasks reference sensitive circumstances---intimate partner violence, medical conditions, immigration status, religious practice---because these are precisely where an agent's failure to apply what it knows carries the greatest cost; excluding them would measure the phenomenon where it matters least. Tasks describe such circumstances only as far as needed to state a user fact.

On dual use: we document a reproducible failure mode across evaluated research and open-source memory systems, including which queries evade retrieval. The same failure could matter in deployed assistants, where it would degrade safety silently and remain difficult for users to detect by the natural means of asking what the agent remembers. Disclosure helps system builders test for the failure without granting an adversary a new capability.

Our results should not be read as endorsing always-in-state memory for deployment: holding a full user profile in context every turn broadens privacy exposure relative to retrieving a narrow slice on demand. Section~\ref{sec:always-in-state} treats it as a diagnostic, not a recommendation.

\section*{Reproducibility Statement}

The construction pipeline is specified in Section~\ref{sec:bench}, the source list in Appendix~\ref{app:source-collections}, and the verbatim extraction prompt in Appendix~\ref{app:generation-prompt}. Both judge prompts appear in Appendix~\ref{app:judge-prompts}; the binary rubric is stated in Section~\ref{sec:eval-setting}. For evaluation, we insert each generated memory turn into an LME-s conversation, continue the simulated dialogue while the memory system operates, and then issue the task queries; Appendix~\ref{app:injection-protocol} gives the exact construction and schedule. Per-system hyperparameters (internal models, embeddings, retrieval depths, search budgets) are itemized in Appendix~\ref{app:system-hparams}. The always-in-state baseline is fully specified by Algorithm~\ref{alg:always-in-state} and its prompts, and needs nothing beyond an LLM and a text file. The benchmark, per-task outputs, judge verdicts, and harness will be released publicly.

\bibliographystyle{plainnat}
\bibliography{references}

\begin{thebibliography}{81}
\providecommand{\natexlab}[1]{#1}
\providecommand{\url}[1]{\texttt{#1}}
\expandafter\ifx\csname urlstyle\endcsname\relax
  \providecommand{\doi}[1]{doi: #1}\else
  \providecommand{\doi}{doi: \begingroup \urlstyle{rm}\Url}\fi

\bibitem[{Alphabet Inc.}(2024)]{alphabetCodeOfConduct}
{Alphabet Inc.}
\newblock Google code of conduct, 2024.
\newblock
  \url{https://abc.xyz/investor/board-and-governance/google-code-of-conduct/}.
\newblock Accessed July 2026.

\bibitem[{American Society for the Prevention of Cruelty to
  Animals}(n.d.)]{aspcaLilies}
{American Society for the Prevention of Cruelty to Animals}.
\newblock Which lilies are toxic to pets?, n.d.
\newblock \url{https://www.aspca.org/news/which-lilies-are-toxic-pets}.
\newblock Accessed July 2026.

\bibitem[{Anthropic}(2026{\natexlab{a}})]{anthropic2026guidance}
{Anthropic}.
\newblock How people ask claude for personal guidance.
\newblock Anthropic Research, 2026{\natexlab{a}}.
\newblock \url{https://www.anthropic.com/research/claude-personal-guidance}.

\bibitem[{Anthropic}(2026{\natexlab{b}})]{anthropic2026opus48}
{Anthropic}.
\newblock Claude opus 4.8 system card, 2026{\natexlab{b}}.
\newblock \url{https://www.anthropic.com/system-cards}.

\bibitem[{Anthropic}(2026{\natexlab{c}})]{anthropic2026sonnet46}
{Anthropic}.
\newblock Claude sonnet 4.6 system card, 2026{\natexlab{c}}.
\newblock \url{https://www.anthropic.com/system-cards}.

\bibitem[Asai et~al.(2023)Asai, Wu, Wang, Sil, and Hajishirzi]{asai2023selfrag}
Akari Asai, Zeqiu Wu, Yizhong Wang, Avirup Sil, and Hannaneh Hajishirzi.
\newblock Self-rag: Learning to retrieve, generate, and critique through
  self-reflection.
\newblock \emph{arXiv preprint arXiv:2310.11511}, 2023.

\bibitem[Borgeaud et~al.(2022)Borgeaud, Mensch, Hoffmann, Cai, Rutherford,
  Millican, van~den Driessche, Lespiau, Damoc, Clark,
  et~al.]{borgeaud2022retro}
Sebastian Borgeaud, Arthur Mensch, Jordan Hoffmann, Trevor Cai, Eliza
  Rutherford, Katie Millican, George~Bm van~den Driessche, Jean-Baptiste
  Lespiau, Bogdan Damoc, Aidan Clark, et~al.
\newblock Improving language models by retrieving from trillions of tokens.
\newblock In \emph{Proceedings of the 39th International Conference on Machine
  Learning}, volume 162 of \emph{Proceedings of Machine Learning Research},
  pages 2206--2240, 2022.
\newblock \url{https://proceedings.mlr.press/v162/borgeaud22a.html}.

\bibitem[{Centers for Disease Control and Prevention}(2026)]{cdcChokingHazards}
{Centers for Disease Control and Prevention}.
\newblock Choking hazards, 2026.
\newblock
  \url{https://www.cdc.gov/infant-toddler-nutrition/foods-and-drinks/choking-hazards.html}.
\newblock Accessed July 2026.

\bibitem[{Centers for Disease Control and
  Prevention}(n.d.{\natexlab{a}})]{cdcIPV}
{Centers for Disease Control and Prevention}.
\newblock About intimate partner violence, n.d.{\natexlab{a}}.
\newblock \url{https://www.cdc.gov/intimate-partner-violence/about/index.html}.
\newblock Accessed July 2026.

\bibitem[{Centers for Disease Control and
  Prevention}(n.d.{\natexlab{b}})]{cdcPregnancyFoodSafety}
{Centers for Disease Control and Prevention}.
\newblock Safer food choices for pregnant people, n.d.{\natexlab{b}}.
\newblock \url{https://www.cdc.gov/food-safety/foods/pregnant-women.html}.
\newblock Accessed July 2026.

\bibitem[Chhikara et~al.(2025)Chhikara, Khant, Aryan, Singh, and
  Yadav]{chadha2025mem0}
Prateek Chhikara, Dev Khant, Saket Aryan, Taranjeet Singh, and Deshraj Yadav.
\newblock Mem0: Building production-ready ai agents with scalable long-term
  memory.
\newblock \emph{arXiv preprint arXiv:2504.19413}, 2025.

\bibitem[{Consumer Financial Protection Bureau}(n.d.)]{cfpbAsk}
{Consumer Financial Protection Bureau}.
\newblock Ask cfpb, n.d.
\newblock \url{https://www.consumerfinance.gov/ask-cfpb/}.
\newblock Accessed July 2026.

\bibitem[{DeepSeek-AI}(2026)]{deepseek2026v4}
{DeepSeek-AI}.
\newblock {DeepSeek-V4} preview release, 2026.
\newblock \url{https://api-docs.deepseek.com/news/news260424/}.

\bibitem[Du et~al.(2026)Du, Xu, Zhu, Wang, Wang, Wang, and Mao]{du2026arag}
Mingxuan Du, Benfeng Xu, Chiwei Zhu, Shaohan Wang, Pengyu Wang, Xiaorui Wang,
  and Zhendong Mao.
\newblock A-rag: Scaling agentic retrieval-augmented generation via
  hierarchical retrieval interfaces, 2026.
\newblock \url{https://arxiv.org/abs/2602.03442}.

\bibitem[Edge et~al.(2024)Edge, Trinh, Cheng, Bradley, Chao, Mody, Truitt,
  Metropolitansky, Ness, and Larson]{edge2024graphrag}
Darren Edge, Ha~Trinh, Newman Cheng, Joshua Bradley, Alex Chao, Apurva Mody,
  Steven Truitt, Dasha Metropolitansky, Robert~Osazuwa Ness, and Jonathan
  Larson.
\newblock From local to global: A graph rag approach to query-focused
  summarization.
\newblock \emph{arXiv preprint arXiv:2404.16130}, 2024.

\bibitem[{Federal Aviation Administration}(2024)]{faaMedications}
{Federal Aviation Administration}.
\newblock Medications and flying, 2024.
\newblock \url{https://www.faa.gov/pilots/medical_certification/medications}.
\newblock Accessed July 2026.

\bibitem[{Federal Trade Commission}(n.d.)]{ftcGiftCardScams}
{Federal Trade Commission}.
\newblock Avoiding and reporting gift card scams, n.d.
\newblock
  \url{https://consumer.ftc.gov/articles/avoiding-and-reporting-gift-card-scams}.
\newblock Accessed July 2026.

\bibitem[{Gemini Team, Google DeepMind}(2025)]{google2025gemini25}
{Gemini Team, Google DeepMind}.
\newblock Gemini 2.5: Pushing the frontier with advanced reasoning,
  multimodality, long context, and next generation agentic capabilities.
\newblock \emph{arXiv preprint arXiv:2507.06261}, 2025.

\bibitem[Gutierrez et~al.(2024)Gutierrez, Shu, Gu, Yasunaga, and
  Su]{gutierrez2024hipporag}
Bernal~Jimenez Gutierrez, Yiheng Shu, Yu~Gu, Michihiro Yasunaga, and Yu~Su.
\newblock Hipporag: Neurobiologically inspired long-term memory for large
  language models.
\newblock \emph{arXiv preprint arXiv:2405.14831}, 2024.

\bibitem[Gutierrez et~al.(2025)Gutierrez, Shu, Qi, Zhou, and
  Su]{gutierrez2025hipporag2}
Bernal~Jimenez Gutierrez, Yiheng Shu, Weijian Qi, Sizhe Zhou, and Yu~Su.
\newblock From rag to memory: Non-parametric continual learning for large
  language models.
\newblock \emph{arXiv preprint arXiv:2502.14802}, 2025.

\bibitem[Hu et~al.(2025)Hu, Wang, and McAuley]{hu2025mabench}
Yuanzhe Hu, Yu~Wang, and Julian McAuley.
\newblock Evaluating memory in llm agents via incremental multi-turn
  interactions.
\newblock \emph{arXiv preprint arXiv:2507.05257}, 2025.

\bibitem[Hu et~al.(2026)Hu, Zhu, Zhao, Liang, Yan, He, and
  Gui]{wang2025xmemory}
Zhanghao Hu, Qinglin Zhu, Runcong Zhao, Di~Liang, Hanqi Yan, Yulan He, and Lin
  Gui.
\newblock Beyond rag for agent memory: Retrieval by decoupling and aggregation.
\newblock \emph{arXiv preprint arXiv:2602.02007}, 2026.

\bibitem[{Internal Revenue Service}(2025)]{irsEarlyDistribution}
{Internal Revenue Service}.
\newblock Exceptions to tax on early distributions, 2025.
\newblock
  \url{https://www.irs.gov/retirement-plans/plan-participant-employee/retirement-topics-exceptions-to-tax-on-early-distributions}.
\newblock Accessed July 2026.

\bibitem[Jeong et~al.(2024)Jeong, Baek, Cho, Hwang, and
  Park]{jeong2024adaptiverag}
Soyeong Jeong, Jinheon Baek, Sukmin Cho, Sung~Ju Hwang, and Jong~C. Park.
\newblock Adaptive-rag: Learning to adapt retrieval-augmented large language
  models through question complexity.
\newblock \emph{arXiv preprint arXiv:2403.14403}, 2024.

\bibitem[Johnson et~al.(2021)Johnson, Douze, and J{\'e}gou]{johnson2019faiss}
Jeff Johnson, Matthijs Douze, and Herv{\'e} J{\'e}gou.
\newblock Billion-scale similarity search with {GPU}s.
\newblock \emph{IEEE Transactions on Big Data}, 7\penalty0 (3):\penalty0
  535--547, 2021.
\newblock \doi{10.1109/TBDATA.2019.2921572}.

\bibitem[Kang et~al.(2025)Kang, Ji, Zhao, and Bai]{kang2025memoryos}
Jiazheng Kang, Mingming Ji, Zhe Zhao, and Ting Bai.
\newblock Memory os of ai agent, 2025.
\newblock \url{https://arxiv.org/abs/2506.06326}.

\bibitem[Karpukhin et~al.(2020)Karpukhin, Oguz, Min, Lewis, Wu, Edunov, Chen,
  and Yih]{karpukhin2020dpr}
Vladimir Karpukhin, Barlas Oguz, Sewon Min, Patrick Lewis, Ledell Wu, Sergey
  Edunov, Danqi Chen, and Wen-tau Yih.
\newblock Dense passage retrieval for open-domain question answering.
\newblock \emph{arXiv preprint arXiv:2004.04906}, 2020.

\bibitem[{King Arthur Baking Company}(n.d.)]{kingArthurMacarons}
{King Arthur Baking Company}.
\newblock Simple macarons recipe, n.d.
\newblock
  \url{https://www.kingarthurbaking.com/recipes/simple-macarons-recipe}.
\newblock Accessed July 2026.

\bibitem[Kynoch et~al.(2023)Kynoch, Latapie, and van~der
  Sluis]{wang2024recallm}
Brandon Kynoch, Hugo Latapie, and Dwane van~der Sluis.
\newblock Recallm: An adaptable memory mechanism with temporal understanding
  for large language models.
\newblock \emph{arXiv preprint arXiv:2307.02738}, 2023.

\bibitem[Lewis et~al.(2020)Lewis, Perez, Piktus, Petroni, Karpukhin, Goyal,
  Kuttler, Lewis, Yih, Rocktaschel, Riedel, and Kiela]{lewis2020rag}
Patrick Lewis, Ethan Perez, Aleksandra Piktus, Fabio Petroni, Vladimir
  Karpukhin, Naman Goyal, Heinrich Kuttler, Mike Lewis, Wen-tau Yih, Tim
  Rocktaschel, Sebastian Riedel, and Douwe Kiela.
\newblock Retrieval-augmented generation for knowledge-intensive nlp tasks.
\newblock \emph{arXiv preprint arXiv:2005.11401}, 2020.

\bibitem[Li et~al.(2026)Li, Guo, Zhang, Xu, Huang, Liu, Xu, Xu, and
  Liu]{li2026locomoplus}
Yifei Li, Weidong Guo, Lingling Zhang, Rongman Xu, Muye Huang, Hui Liu, Lijiao
  Xu, Yu~Xu, and Jun Liu.
\newblock Locomo-plus: Beyond-factual cognitive memory evaluation framework for
  llm agents.
\newblock \emph{arXiv preprint arXiv:2602.10715}, 2026.

\bibitem[Liang et~al.(2024)Liang, He, Xia, Song, Wang, Tao,
  et~al.]{wang2024sage}
Xuechen Liang, Yangfan He, Yinghui Xia, Xinyuan Song, Jianhui Wang, Meiling
  Tao, et~al.
\newblock Self-evolving agents with reflective and memory-augmented abilities.
\newblock \emph{arXiv preprint arXiv:2409.00872}, 2024.

\bibitem[{Love is Respect}(n.d.{\natexlab{a}})]{loveisrespectDigital}
{Love is Respect}.
\newblock Digital boundaries, n.d.{\natexlab{a}}.
\newblock \url{https://www.loveisrespect.org/resources/digital-boundaries/}.
\newblock Accessed July 2026.

\bibitem[{Love is Respect}(n.d.{\natexlab{b}})]{loveisrespectPasswords}
{Love is Respect}.
\newblock Should we share passwords?, n.d.{\natexlab{b}}.
\newblock
  \url{https://www.loveisrespect.org/resources/should-we-share-passwords/}.
\newblock Accessed July 2026.

\bibitem[Maharana et~al.(2024)Maharana, Lee, Tulyakov, Bansal, Barbieri, and
  Fang]{maharana2024locomo}
Adyasha Maharana, Dong-Ho Lee, Sergey Tulyakov, Mohit Bansal, Francesco
  Barbieri, and Yuwei Fang.
\newblock Evaluating very long-term conversational memory of llm agents.
\newblock \emph{arXiv preprint arXiv:2402.17753}, 2024.

\bibitem[{MedlinePlus}(n.d.)]{medlinePlusHealthTopics}
{MedlinePlus}.
\newblock Health topics, n.d.
\newblock \url{https://medlineplus.gov/healthtopics.html}.
\newblock Accessed July 2026.

\bibitem[{National Center for O*NET Development}(n.d.)]{onetDatabase}
{National Center for O*NET Development}.
\newblock O*net database, n.d.
\newblock \url{https://www.onetcenter.org/database.html}.
\newblock Accessed July 2026.

\bibitem[{National Domestic Violence
  Hotline}(n.d.{\natexlab{a}})]{hotlineIdentifyAbuse}
{National Domestic Violence Hotline}.
\newblock Identify abuse, n.d.{\natexlab{a}}.
\newblock \url{https://www.thehotline.org/identify-abuse}.
\newblock Accessed July 2026.

\bibitem[{National Domestic Violence
  Hotline}(n.d.{\natexlab{b}})]{hotlineInternetSafety}
{National Domestic Violence Hotline}.
\newblock Internet safety for survivors, n.d.{\natexlab{b}}.
\newblock \url{https://www.thehotline.org/plan-for-safety/internet-safety/}.
\newblock Accessed July 2026.

\bibitem[{National Domestic Violence
  Hotline}(n.d.{\natexlab{c}})]{hotlineTechAbuse}
{National Domestic Violence Hotline}.
\newblock Technology-facilitated abuse, n.d.{\natexlab{c}}.
\newblock
  \url{https://www.thehotline.org/resources/technology-facilitated-abuse/}.
\newblock Accessed July 2026.

\bibitem[{Novartis Pharmaceuticals Corporation}(n.d.)]{fdaTegretol}
{Novartis Pharmaceuticals Corporation}.
\newblock Tegretol (carbamazepine) prescribing information, n.d.
\newblock
  \url{https://dailymed.nlm.nih.gov/dailymed/lookup.cfm?setid=8d409411-aa9f-4f3a-a52c-fbcb0c3ec053}.
\newblock FDA-approved labeling via DailyMed; accessed July 2026.

\bibitem[{Occupational Safety and Health
  Administration}(n.d.{\natexlab{a}})]{oshaFacialHair}
{Occupational Safety and Health Administration}.
\newblock Respiratory protection and facial hair, n.d.{\natexlab{a}}.
\newblock \url{https://www.osha.gov/node/34013}.
\newblock Accessed July 2026.

\bibitem[{Occupational Safety and Health
  Administration}(n.d.{\natexlab{b}})]{oshaHeat}
{Occupational Safety and Health Administration}.
\newblock Heat exposure, n.d.{\natexlab{b}}.
\newblock \url{https://www.osha.gov/heat-exposure}.
\newblock Accessed July 2026.

\bibitem[{Occupational Safety and Health
  Administration}(n.d.{\natexlab{c}})]{oshaNoise}
{Occupational Safety and Health Administration}.
\newblock Occupational noise exposure, n.d.{\natexlab{c}}.
\newblock \url{https://www.osha.gov/noise}.
\newblock Accessed July 2026.

\bibitem[{Occupational Safety and Health
  Administration}(n.d.{\natexlab{d}})]{oshaRespiratoryProtection}
{Occupational Safety and Health Administration}.
\newblock Respiratory protection, n.d.{\natexlab{d}}.
\newblock \url{https://www.osha.gov/respiratory-protection}.
\newblock Accessed July 2026.

\bibitem[{Open Food Facts}(n.d.)]{openFoodFacts}
{Open Food Facts}.
\newblock Open food facts product database, n.d.
\newblock \url{https://world.openfoodfacts.org/data}.
\newblock Accessed July 2026.

\bibitem[{OpenAI}(2024{\natexlab{a}})]{openai2024embeddings}
{OpenAI}.
\newblock New embedding models and {API} updates, 2024{\natexlab{a}}.
\newblock \url{https://openai.com/index/new-embedding-models-and-api-updates/}.

\bibitem[{OpenAI}(2024{\natexlab{b}})]{openai2024gpt4o}
{OpenAI}.
\newblock {GPT-4o} system card, 2024{\natexlab{b}}.
\newblock \url{https://openai.com/index/gpt-4o-system-card/}.

\bibitem[{OpenAI}(2025)]{openai2025gpt5}
{OpenAI}.
\newblock {GPT-5} system card, 2025.
\newblock \url{https://openai.com/index/gpt-5-system-card/}.

\bibitem[Packer et~al.(2023)Packer, Wooders, Lin, Fang, Patil, Stoica, and
  Gonzalez]{packer2023memgpt}
Charles Packer, Sarah Wooders, Kevin Lin, Vivian Fang, Shishir~G. Patil, Ion
  Stoica, and Joseph~E. Gonzalez.
\newblock Memgpt: Towards llms as operating systems.
\newblock \emph{arXiv preprint arXiv:2310.08560}, 2023.

\bibitem[Pan et~al.(2025)Pan, Wu, Jiang, Luo, Cheng, Li, Yang, Lin, Zhao, Qiu,
  and Gao]{liu2025secom}
Zhuoshi Pan, Qianhui Wu, Huiqiang Jiang, Xufang Luo, Hao Cheng, Dongsheng Li,
  Yuqing Yang, Chin-Yew Lin, H.~Vicky Zhao, Lili Qiu, and Jianfeng Gao.
\newblock On memory construction and retrieval for personalized conversational
  agents.
\newblock \emph{arXiv preprint arXiv:2502.05589}, 2025.

\bibitem[Park et~al.(2023)Park, O'Brien, Cai, Morris, Liang, and
  Bernstein]{park2023generative}
Joon~Sung Park, Joseph~C. O'Brien, Carrie~J. Cai, Meredith~Ringel Morris, Percy
  Liang, and Michael~S. Bernstein.
\newblock Generative agents: Interactive simulacra of human behavior.
\newblock \emph{arXiv preprint arXiv:2304.03442}, 2023.

\bibitem[Qin et~al.(2026)Qin, Feng, Ma, Feng, and
  Kong]{qin2026implicitmembench}
Chonghan Qin, Xiachong Feng, Weitao Ma, Xiaocheng Feng, and Lingpeng Kong.
\newblock Implicitmembench: Measuring unconscious behavioral adaptation in
  large language models.
\newblock In \emph{Proceedings of the 64th Annual Meeting of the Association
  for Computational Linguistics}, pages 28232--28261, 2026.

\bibitem[Rasmussen et~al.(2025)Rasmussen, Paliychuk, Beauvais, Ryan, and
  Chalef]{strang2025zep}
Preston Rasmussen, Pavlo Paliychuk, Travis Beauvais, Jack Ryan, and Daniel
  Chalef.
\newblock Zep: A temporal knowledge graph architecture for agent memory.
\newblock \emph{arXiv preprint arXiv:2501.13956}, 2025.

\bibitem[Reimers and Gurevych(2019)]{reimers2019sentencebert}
Nils Reimers and Iryna Gurevych.
\newblock Sentence-{BERT}: Sentence embeddings using siamese {BERT}-networks.
\newblock In \emph{Proceedings of the 2019 Conference on Empirical Methods in
  Natural Language Processing and the 9th International Joint Conference on
  Natural Language Processing}, pages 3982--3992, 2019.

\bibitem[Robertson and Zaragoza(2009)]{robertson2009bm25}
Stephen Robertson and Hugo Zaragoza.
\newblock The probabilistic relevance framework: {BM25} and beyond.
\newblock \emph{Foundations and Trends in Information Retrieval}, 3\penalty0
  (4):\penalty0 333--389, 2009.
\newblock \doi{10.1561/1500000019}.

\bibitem[Salama et~al.(2025)Salama, Cai, Yuan, Currey, Sunkara, Zhang, and
  Benajiba]{chen2025meminsight}
Rana Salama, Jason Cai, Michelle Yuan, Anna Currey, Monica Sunkara, Yi~Zhang,
  and Yassine Benajiba.
\newblock Meminsight: Autonomous memory augmentation for llm agents.
\newblock \emph{arXiv preprint arXiv:2503.21760}, 2025.

\bibitem[Sarthi et~al.(2024)Sarthi, Abdullah, Tuli, Khanna, Goldie, and
  Manning]{sarthi2024raptor}
Parth Sarthi, Salman Abdullah, Aditi Tuli, Shubh Khanna, Anna Goldie, and
  Christopher~D. Manning.
\newblock Raptor: Recursive abstractive processing for tree-organized
  retrieval.
\newblock \emph{arXiv preprint arXiv:2401.18059}, 2024.

\bibitem[{Substance Abuse and Mental Health Services
  Administration}(n.d.)]{samhsaRecovery}
{Substance Abuse and Mental Health Services Administration}.
\newblock Recovery and recovery support, n.d.
\newblock \url{https://www.samhsa.gov/find-help/recovery}.
\newblock Accessed July 2026.

\bibitem[{Tesla}(n.d.)]{teslaSupercharging}
{Tesla}.
\newblock Supercharging, n.d.
\newblock \url{https://www.tesla.com/support/charging/supercharging}.
\newblock Accessed July 2026.

\bibitem[{U.S. Citizenship and Immigration
  Services}(n.d.)]{uscisTravelDocuments}
{U.S. Citizenship and Immigration Services}.
\newblock Travel documents, n.d.
\newblock
  \url{https://www.uscis.gov/green-card/green-card-processes-and-procedures/travel-documents}.
\newblock Accessed July 2026.

\bibitem[{U.S. Consumer Product Safety Commission}(2026)]{cpscUhomeproRecall}
{U.S. Consumer Product Safety Commission}.
\newblock Recall no. 26448: Uhomepro 5-drawer dressers, 2026.
\newblock
  \url{https://www.saferproducts.gov/RestWebServices/Recall?RecallID=10745\&format=json}.
\newblock Accessed July 2026.

\bibitem[{U.S. Consumer Product Safety
  Commission}(n.d.{\natexlab{a}})]{cpscRecalls}
{U.S. Consumer Product Safety Commission}.
\newblock Recalls, n.d.{\natexlab{a}}.
\newblock \url{https://www.cpsc.gov/Recalls}.
\newblock Accessed July 2026.

\bibitem[{U.S. Consumer Product Safety
  Commission}(n.d.{\natexlab{b}})]{cpscSmallParts}
{U.S. Consumer Product Safety Commission}.
\newblock Small parts and choking hazard labeling faqs, n.d.{\natexlab{b}}.
\newblock
  \url{https://www.cpsc.gov/FAQ/Small-Parts-and-Choking-Hazard-Labeling-FAQs}.
\newblock Accessed July 2026.

\bibitem[{U.S. Department of Health and Human
  Services}(n.d.)]{hhsPhysicalActivity}
{U.S. Department of Health and Human Services}.
\newblock Physical activity guidelines for americans, n.d.
\newblock
  \url{https://odphp.health.gov/our-work/nutrition-physical-activity/physical-activity-guidelines/current-guidelines}.
\newblock Accessed July 2026.

\bibitem[{U.S. Department of Labor}(n.d.)]{dolFmla}
{U.S. Department of Labor}.
\newblock Family and medical leave act frequently asked questions, n.d.
\newblock \url{https://www.dol.gov/agencies/whd/fmla/faq}.
\newblock Accessed July 2026.

\bibitem[{U.S. Food and Drug Administration}(n.d.)]{fdaFoodAllergies}
{U.S. Food and Drug Administration}.
\newblock Food allergies, n.d.
\newblock
  \url{https://www.fda.gov/food/nutrition-food-labeling-and-critical-foods/food-allergies}.
\newblock Accessed July 2026.

\bibitem[{U.S. National Library of Medicine}(n.d.)]{dailyMed}
{U.S. National Library of Medicine}.
\newblock Dailymed: All drug labels, n.d.
\newblock
  \url{https://dailymed.nlm.nih.gov/dailymed/spl-resources-all-drug-labels.cfm}.
\newblock Accessed July 2026.

\bibitem[Wang et~al.(2023)Wang, Xie, Jiang, Mandlekar, Xiao, Zhu, Fan, and
  Anandkumar]{wang2023voyager}
Guanzhi Wang, Yuqi Xie, Yunfan Jiang, Ajay Mandlekar, Chaowei Xiao, Yuke Zhu,
  Linxi Fan, and Anima Anandkumar.
\newblock Voyager: An open-ended embodied agent with large language models.
\newblock \emph{arXiv preprint arXiv:2305.16291}, 2023.

\bibitem[Wang et~al.(2020)Wang, Wei, Dong, Bao, Yang, and Zhou]{wang2020minilm}
Wenhui Wang, Furu Wei, Li~Dong, Hangbo Bao, Nan Yang, and Ming Zhou.
\newblock {MiniLM}: Deep self-attention distillation for task-agnostic
  compression of pre-trained transformers.
\newblock In \emph{Advances in Neural Information Processing Systems},
  volume~33, pages 5776--5788, 2020.

\bibitem[{Wikipedia contributors}(n.d.{\natexlab{a}})]{wikipediaGelatin}
{Wikipedia contributors}.
\newblock Gelatin, n.d.{\natexlab{a}}.
\newblock \url{https://en.wikipedia.org/wiki/Gelatin}.
\newblock Accessed July 2026.

\bibitem[{Wikipedia
  contributors}(n.d.{\natexlab{b}})]{wikipediaIslamicDietaryLaws}
{Wikipedia contributors}.
\newblock Islamic dietary laws, n.d.{\natexlab{b}}.
\newblock \url{https://en.wikipedia.org/wiki/Islamic_dietary_laws}.
\newblock Accessed July 2026.

\bibitem[{World Wide Web
  Consortium}(2025{\natexlab{a}})]{w3cLabelsInstructions}
{World Wide Web Consortium}.
\newblock Understanding success criterion 3.3.2: Labels or instructions,
  2025{\natexlab{a}}.
\newblock
  \url{https://www.w3.org/WAI/WCAG21/Understanding/labels-or-instructions.html}.
\newblock Accessed July 2026.

\bibitem[{World Wide Web Consortium}(2025{\natexlab{b}})]{w3cUseOfColor}
{World Wide Web Consortium}.
\newblock Understanding success criterion 1.4.1: Use of color,
  2025{\natexlab{b}}.
\newblock \url{https://www.w3.org/WAI/WCAG21/Understanding/use-of-color.html}.
\newblock Accessed July 2026.

\bibitem[Wu et~al.(2024)Wu, Wang, Yu, Zhang, Chang, and Yu]{wu2024longmemeval}
Di~Wu, Hongwei Wang, Wenhao Yu, Yuwei Zhang, Kai-Wei Chang, and Dong Yu.
\newblock Longmemeval: Benchmarking chat assistants on long-term interactive
  memory.
\newblock \emph{arXiv preprint arXiv:2410.10813}, 2024.

\bibitem[Xiao et~al.(2024)Xiao, Liu, Zhang, Muennighoff, Lian, and
  Nie]{xiao2023cpack}
Shitao Xiao, Zheng Liu, Peitian Zhang, Niklas Muennighoff, Defu Lian, and
  Jian-Yun Nie.
\newblock C-pack: Packed resources for general chinese embeddings.
\newblock In \emph{Proceedings of the 47th International ACM SIGIR Conference
  on Research and Development in Information Retrieval}, 2024.

\bibitem[Xu et~al.(2025)Xu, Liang, Mei, Gao, Tan, and Zhang]{xu2025amem}
Wujiang Xu, Zujie Liang, Kai Mei, Hang Gao, Juntao Tan, and Yongfeng Zhang.
\newblock A-mem: Agentic memory for llm agents.
\newblock \emph{arXiv preprint arXiv:2502.12110}, 2025.

\bibitem[Yan et~al.(2025)Yan, Li, Qian, Lu, and Liu]{yan2025gam}
B.~Y. Yan, Chaofan Li, Hongjin Qian, Shuqi Lu, and Zheng Liu.
\newblock General agentic memory via deep research, 2025.
\newblock \url{https://arxiv.org/abs/2511.18423}.

\bibitem[Yan et~al.(2024)Yan, Gu, Zhu, and Ling]{yan2024crag}
Shi-Qi Yan, Jia-Chen Gu, Yun Zhu, and Zhen-Hua Ling.
\newblock Corrective retrieval augmented generation.
\newblock \emph{arXiv preprint arXiv:2401.15884}, 2024.

\bibitem[Zhang et~al.(2024)Zhang, Bo, Ma, Li, Chen, Dai, Zhu, Dong, and
  Wen]{zhang2024memorymechanism}
Zeyu Zhang, Xiaohe Bo, Chen Ma, Rui Li, Xu~Chen, Quanyu Dai, Jieming Zhu,
  Zhenhua Dong, and Ji-Rong Wen.
\newblock A survey on the memory mechanism of large language model based
  agents.
\newblock \emph{arXiv preprint arXiv:2404.13501}, 2024.

\bibitem[Zhong et~al.(2024)Zhong, Guo, Gao, Ye, and Wang]{zhong2024memorybank}
Wanjun Zhong, Lianghong Guo, Qiqi Gao, He~Ye, and Yanlin Wang.
\newblock Memorybank: Enhancing large language models with long-term memory.
\newblock \emph{Proceedings of the AAAI Conference on Artificial Intelligence},
  38\penalty0 (17):\penalty0 19724--19731, 2024.
\newblock \doi{10.1609/aaai.v38i17.29946}.

\end{thebibliography}

\clearpage
\beginappendix
\section{Content-Filter Details}
\label{app:filter-details}

Section~\ref{sec:filtering} summarizes the three filters that reduce 1,000 extracted candidates to the evaluation set. We give each rule in full here. Every filter is a function of task content and the fixed background trace alone; none inspects any system's output, and the task set was fixed before scoring.

\paragraph{(i) Similarity filter.}
Each candidate supplies a target memory (the \texttt{user\_message}) and an indirect \texttt{query}. We score the pair two ways: BM25 for lexical overlap, and cosine similarity between \texttt{all-MiniLM-L6-v2} embeddings for dense semantic proximity. A candidate is discarded when either score marks the target as an obvious match to its own query, because such a pair supplies exactly the retrieval cue the benchmark is meant to withhold: a system could surface the memory by similarity alone, without applying any world knowledge, so the task would not test implicit association. This step retains 300 of the 1,000 candidates. Figure~\ref{fig:similarity} is a held-out check rather than a report from the filtering model: it recomputes all four distributions with BGE-small-en-v1.5, using the retrieval prefix recommended by its authors, and compares the retained tasks with LoCoMo, LoCoMo-Plus, and LME-s turns.

\paragraph{(ii) Conflict filter.}
Every task is injected into the same 47-session LME-s background trace (Appendix~\ref{app:injection-protocol}), which already establishes a persona: the speaker's occupation, household, habits, and history. An injected user fact that contradicts this persona---claiming a cat in a trace where the user states they have no pets, or a pilot's medical exam for a speaker established as a nurse---yields a task that is incoherent rather than difficult, and a system could fail it for reasons unrelated to the blind spot. An LLM judge compares each candidate's user fact against the trace and discards contradictions. This filter depends only on the task and the fixed trace, both of which are identical for every system evaluated.

\paragraph{(iii) Expert verification.}
Human experts review each surviving candidate against its source chunk and check four properties: the bridge is factually correct and actually supported by the cited source; the user fact materially changes the correct answer, rather than merely being mentionable; the two turns read as natural conversation; and the indirect query retains no overt cue to the fact, in particular no synonym or same-topic reference that leaked past the automatic filter. Reviewers may retain a candidate, revise it---most often by rewriting an indirect query that disclosed its own answer---or reject it. Experts also confirm that the association is genuinely strong rather than merely arguable, since a bridge a careful reader would dispute cannot support a binary judgment.

\section{Memory-Benchmark Taxonomy}
\label{app:benchmark-taxonomy}

\begin{figure}[H]
  \centering
  \includegraphics[width=0.92\textwidth]{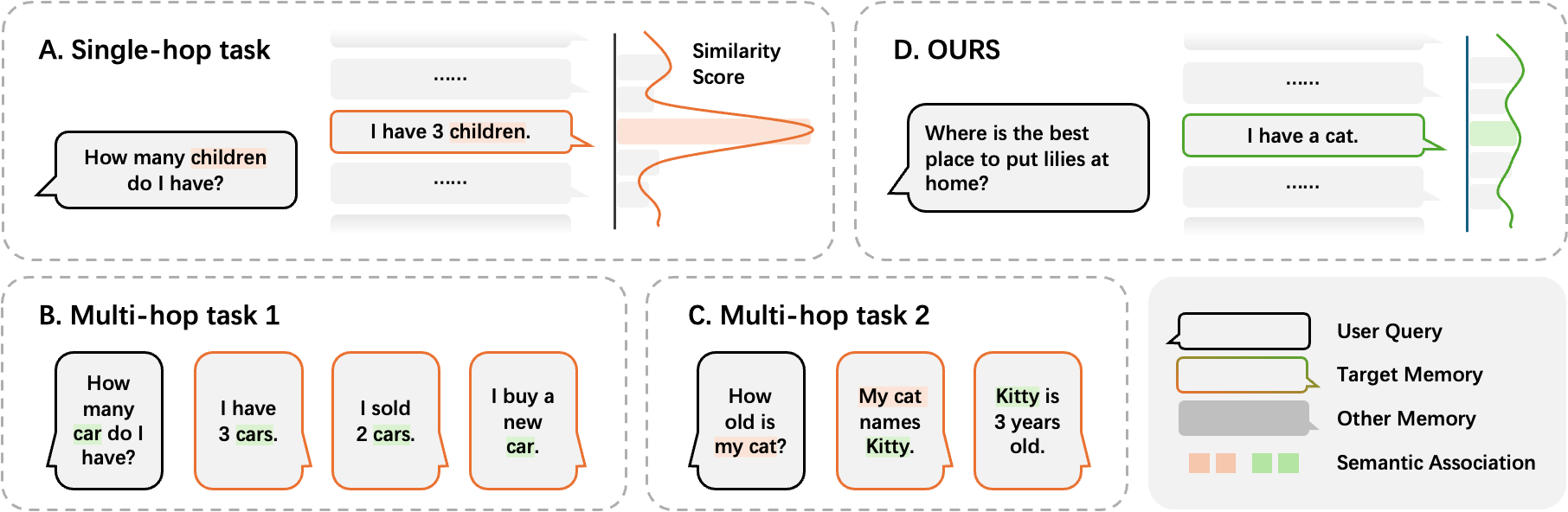}
  \caption{Memory-evaluation settings: explicit recall, state aggregation, entity-based multi-hop reasoning, and InMind's knowledge-mediated implicit application.}
  \label{fig:memory-benchmark-taxonomy}
\end{figure}

\section{Representative Tasks by Domain}
\label{app:domain-examples}

To illustrate the benchmark's coverage, we show one retained task from each domain. Each example gives the stored user memory, the later indirect query, and the knowledge-mediated bridge that should change the answer.

\newcommand{\domainexample}[3]{\begin{quote}\small
\textit{Memory:} #1\par
\textit{Query:} #2\par
\textit{Bridge:} #3
\end{quote}}

\paragraph{Health and wellness (Task 155).}
\domainexample{``I have a tree nut allergy.''}{``I want to try making macarons this weekend. Any good recipes?''}{Standard macarons use almond flour, so the assistant should warn that making or eating them could trigger the user's allergy \citep{kingArthurMacarons,fdaFoodAllergies}.}

\paragraph{Professional and career (Task 4).}
\domainexample{``I'm preparing for my pilot medical exam.''}{``I've been experiencing allergies lately. Can I take antihistamines?''}{Antihistamines can impair flight safety and may entail no-fly periods; the answer should advise checking with an Aviation Medical Examiner \citep{faaMedications}.}

\paragraph{Relationships (Task 113).}
\domainexample{``I am currently experiencing domestic violence.''}{``I've heard there are places that offer temporary accommodation. Do you have any information?''}{The assistant should connect temporary accommodation to domestic-violence shelters and safety resources \citep{hotlineIdentifyAbuse}.}

\paragraph{Financial (Task 213).}
\domainexample{``I just accepted an offer at Google.''}{``I'm thinking of buying some Apple stock. Is it a good investment right now?''}{The answer should note that an investment in a competitor can create, or appear to create, a conflict depending on the user's role, and should point the user to the applicable company policy \citep{alphabetCodeOfConduct}.}

\paragraph{Personal development (Task 121).}
\domainexample{``I have some cognitive impairment and find complex information difficult to process.''}{``I'm registering for an account on a new website. Any advice for setting up a username?''}{The assistant should recommend checking clear character and format rules, and using a simple, easily verified entry process \citep{w3cLabelsInstructions}.}

\paragraph{Spirituality (Task 124).}
\domainexample{``I follow Halal dietary rules.''}{``I need to take a new medicine in capsule form. Is there anything special about the capsules?''}{Gelatin capsules can have animal-derived ingredients, so the answer should suggest verifying that their source meets Halal requirements \citep{wikipediaGelatin}.}

\paragraph{Legal (Task 40).}
\domainexample{``My I-485 application is pending with USCIS.''}{``I'm planning a trip to Canada for a few days. Is there anything I should be aware of?''}{Leaving the United States without Advance Parole can jeopardize a pending I-485 application; the assistant should flag this before offering routine travel advice \citep{uscisTravelDocuments}.}

\paragraph{Consumer (Task 129).}
\domainexample{``I have three young children at home.''}{``I saw a UHOMEPRO chest of drawers I like. Is it worth buying?''}{The product poses tip-over and entrapment risks to children, so the answer should warn about the specific recall and direct the user to its safety and refund instructions \citep{cpscUhomeproRecall}.}

\paragraph{Parenting (Task 13).}
\domainexample{``I have a two-year-old child.''}{``How can I make corn kernels taste good?''}{Whole corn kernels can be a choking risk for toddlers; preparation advice should include cutting or mashing them \citep{cdcChokingHazards}.}

\paragraph{Other (Task 135).}
\domainexample{``I drive a Tesla Model 3.''}{``I want to take a road trip through rural Montana and Wyoming. What routes do you recommend?''}{EV range and up-to-date Supercharger locations should shape the route; Tesla's Trip Planner adds charging stops based on the route and current state of charge \citep{teslaSupercharging}.}

\section{Representative System Responses}
\label{app:response-examples}

We show the complete recorded Task-155 responses for the target-turn Backbone control and xMemory with text-embedding-3-large. The Backbone control applies the in-context allergy correctly. xMemory answers the naive query correctly, but its indirect-query response is judged incorrect because it does not apply the known tree-nut allergy to the current user.


\paragraph{Backbone control (GPT-5-mini).}
\textbf{Indirect response (target memory in context) (judge score: 1).}
\begin{promptbox}\small\sloppy
Important: classic French macarons are made with almond flour and are unsafe for someone with a tree-nut allergy. Don't use almond meal/almond flour or any product labeled "may contain tree nuts." If your allergy is new or you had hives, you might want to check with your doctor or allergist about how strict you need to be and whether you should carry an epinephrine auto-injector.

If you still want a macaron-style treat this weekend, here are two safe, nut-free approaches - one that mimics the look of macarons using nut-free meringue shells, and one that's an almond-free macaron alternative (with caveats about seed flours).

Option A - Nut-free "macaron-style" meringue sandwich cookies (recommended)
\par
These use egg-white meringue shells (no nut flours) and sandwich fillings (buttercream, ganache, or jam). Texture will be crisp shells with a soft/chewy interior depending on baking time - similar and very tasty.

Ingredients (about 18-24 cookies / 9-12 sandwiches)
\par
- 4 large egg whites, room temperature
\par
- Pinch salt or 1/8 tsp cream of tartar
\par
- 200 g (1 cup) granulated/caster sugar
\par
- 75 g (about 2/3 cup) sifted powdered sugar
\par
- 1 tbsp cornstarch (optional, for extra stability)
\par
- 1 tsp vanilla (or other flavor)
\par
- Food coloring (optional)

Basic buttercream filling
\par
- 113 g (1/2 cup) unsalted butter, room temp
\par
- 140-200 g (1-1 2/3 cups) powdered sugar, to taste
\par
- 1-2 tbsp milk/cream
\par
- 1 tsp vanilla

Method
\par
1. Preheat oven to 275 degreesF (135 degreesC). Line baking sheets with parchment or silicone mats. Fit a piping bag with a 1/2-3/4 inch round tip.
\par
2. Put egg whites and granulated sugar in a heatproof bowl over simmering water (bain-marie). Gently whisk until mixture reaches about 60-70 degreesC (140-160 degreesF) and sugar is dissolved (bowl is warm and smooth).
\par
3. Remove from heat and whip with a stand or hand mixer until glossy stiff peaks form. Add vanilla.
\par
4. Sift powdered sugar and cornstarch over the meringue and gently fold to combine. Add food coloring if you like; be gentle so you don't deflate the meringue.
\par
5. Pipe rounds (about 1½ inch) spaced apart. Tap the tray once to remove big air bubbles. Let the piped rounds sit at room temperature for 20-40 minutes until they form a slight skin (dry to the touch).
\par
6. Bake 12-16 minutes (depending on size). The cookies should lift easily and be dry on the outside; if you want a chewier center, bake a minute or two less. Let cool completely on the sheet.
\par
7. Whip butter and powdered sugar and add milk to desired consistency; sandwich cookies with about 1-2 tsp filling each.

Notes/troubleshooting
\par
- Humidity affects meringues - pick a dry day if possible.  
\par
- If shells crack or flatten, you may have overmixed while folding or not rested long enough.  
\par
- These are nut-free but check all packaged ingredients for "may contain tree nuts" warnings.

Option B - Almond-free "macaron" using seed or grain flours (use with caution)
\par
Some people replace almond flour with blanched sunflower seed flour, pumpkin-seed flour, or finely ground oats. Results vary (color, flavor, texture), and there are two important caveats:
\par
- Seeds (sunflower/pumpkin) can cause allergic cross-reactions in some people with tree-nut allergies - check with your doctor if you're unsure.
\par
- Seed flours can sometimes oxidize and turn green when mixed with egg whites; this is harmless but surprising.

If you want to try this, use a tested recipe that swaps in the seed flour measured and sifted like almond meal and follow standard macaron technique (aging egg whites, macaronage, rest until skins form, bake at \textasciitilde{}300 degreesF/150 degreesC for \textasciitilde{}12-16 min). I can paste a specific seed-flour macaron recipe if you want to experiment - but if your allergy is severe, the meringue sandwich cookies above are the safer nut-free route.

Allergy safety tips
\par
- Read labels: many packaged ingredients may be processed in facilities that handle tree nuts - look for "may contain" warnings and avoid those if you require strict avoidance.  
\par
- Clean any surfaces/utensils that might have touched nuts.  
\par
- If ordering any additional ingredients (e.g., chocolate, puffed grains), buy single-ingredient packages that are labeled nut-free or from a trusted brand.  
\par
- If you're not yet cleared by an allergist to try seeds/coconut, avoid substituting those without guidance.

If you want, I can:
\par
- Send a detailed printable recipe (with weights and step-by-step times) for the nut-free meringue sandwich cookies and a couple of filling flavor ideas; or
\par
- Send an almond-free seed-flour macaron recipe (with green-tinge warning) if you'd like to experiment.

Which would you prefer?
\end{promptbox}

\paragraph{xMemory (emb3-large).}
\textbf{Naive response (judge score: 1).}
\begin{promptbox}\small\sloppy
You told me you have a tree nut allergy - you developed hives after eating trail mix. I also know you're aware this means avoiding tree nuts (eg. almonds, cashews, walnuts) and being careful about hidden ingredients and cross-contamination. If you haven't already, consider talking to a doctor about testing, an action plan, and whether you should carry an epinephrine auto-injector.
\end{promptbox}
\textbf{Indirect response (judge score: 0).}
\begin{promptbox}\small\sloppy
Fantastic - macarons are a lovely (and achievable) weekend project. Below is a reliable recipe using the Italian-meringue method (more forgiving than the French method) plus a mango-white chocolate ganache filling so you can indulge your mango obsession. I'll also include quick troubleshooting tips, storage, and a pregnancy/immunity note.

Yield: \textasciitilde{}20 filled macarons (40 shells)

Ingredients
\par
- Shells (Italian meringue method)
\par
  - 110 g almond flour (finely ground)
\par
  - 200 g powdered (confectioners') sugar
\par
  - 90 g egg whites (about 3 large), room temperature and divided (see method)
\par
  - 150 g granulated sugar
\par
  - 40 g water

- Mango White Chocolate Ganache filling
\par
  - 100 g ripe mango puree (fresh or thawed frozen; strained if fibrous)
\par
  - 200 g good-quality white chocolate, finely chopped
\par
  - 15-20 g unsalted butter (optional - adds shine and stability)
\par
  - Pinch of salt, optional 1 tsp light corn syrup or glucose (for texture)

Equipment and tips
\par
- Digital scale (very helpful - measure by weight)
\par
- Fine mesh sieve or doubled sifter
\par
- Stand mixer or hand mixer
\par
- Candy thermometer (for sugar to 118 degreesC / 244 degreesF)
\par
- Piping bag with round tip (about 8 mm)
\par
- Parchment paper or silicone macaron mat
\par
- Baking sheets (rotate during bake if needed)

Method - shells (Italian meringue)
\par
1. Prep: Line baking sheets. Sift almond flour + powdered sugar together at least twice (removing any lumps). Measure the mixture by weight; set aside.
\par
2. Make a paste: Add 30-35 g of the egg whites to the dry mix and stir to a thick, smooth paste. This helps with macaronage and reduces lumps.
\par
3. Make the Italian meringue: In a small saucepan combine granulated sugar + water and heat to 118 degreesC (244 degreesF). While it cooks, whip the remaining 55-60 g egg whites to soft peaks. When the syrup reaches 118 degreesC, pour it slowly in a thin stream into the whipping whites with the mixer on medium-high. Whip to glossy stiff peaks and let cool a little.
\par
4. Macaronage: Fold about one-third of the meringue into the almond paste to loosen it, then fold in the rest gently. The goal is a "lava" ribbon: batter should fall in thick ribbons and smooth out in \textasciitilde{}10-15 seconds. Stop when the batter flows slowly off the spatula and reaches that stage - overmixing makes flat shells, undermixing gives peaks and no feet.
\par
5. Pipe: Transfer to a piping bag and pipe 3-4 cm (1.25-1.5") rounds about 2.5 cm apart. Tap the tray firmly against the counter 2-3 times to release air bubbles. Pop any visible air bubbles with a toothpick.
\par
6. Rest: Let the piped shells sit at room temperature for 20-60 minutes (depends on humidity) until a thin skin forms - they should not be tacky to the touch.
\par
7. Bake: Preheat oven to 150 degreesC (300 degreesF). Bake 14-18 minutes (start checking at 12-14). Rotate halfway if needed. Shells are ready when they have feet and the tops are set; they should not brown.
\par
8. Cool: Let cool completely on the tray before removing.

Mango white chocolate ganache (filling)
\par
1. Warm the mango puree gently to just under simmer. If it's very watery, simmer briefly to concentrate (don't over-reduce or it will darken).
\par
2. Place chopped white chocolate in a bowl. Pour hot mango puree over chocolate; let sit 1-2 minutes, then stir until smooth and emulsified. Stir in butter and a pinch of salt; add 1 tsp corn syrup if desired for texture.
\par
3. Chill until pipeable (refrigerate 30-60 min). If too stiff, warm gently; if too soft, chill longer.

Assemble and mature
\par
- Pair shells by size. Pipe a small dollop of ganache on one shell and sandwich with its partner.
\par
- For best texture and flavor, refrigerate assembled macarons in an airtight container for 24-48 hours (maturing dramatically improves chew and flavor). Bring to room temperature before serving (20-30 minutes).

Troubleshooting (quick)
\par
- No feet / flat: likely undermixed batter or oven too hot/too cold. Check macaronage and rest time.
\par
- Cracked tops: oven temp too high or shells didn't develop a skin before baking.
\par
- Hollow shells: overbaking or too much heat late in bake; also undermixed batter can create hollows.
\par
- Lopsided peaks/uneven: leftover air bubbles; tap tray thoroughly and pop bubbles.

Safety / special notes
\par
- Italian meringue uses cooked sugar and gives a safer meringue than raw French method; if you're pregnant or immunocompromised, prefer Italian meringue or use pasteurized egg whites.
\par
- Use fresh, room-temp egg whites for best volume. Aging egg whites in the fridge for a day (covered) and bringing back to room temp can help some bakers.

Variations and ideas
\par
- Luxemburgerli-style: pipe smaller (bite-sized) shells and bake just a little less for a softer center.
\par
- Mango curd alternative: make a mango curd (egg-yolk based) for an intense mango flavor, but it needs to be thick/stable and refrigerated. A white-chocolate ganache with mango puree is typically easier and more stable.
\par
- Add lime zest to the ganache for brightness, or fold a little finely chopped freeze-dried mango into the filling for texture.

If you want, I can:
\par
- Give a shorter French-method recipe (no candy thermometer), or
\par
- Walk you through a timeline for your baking day (what to make first, how long to rest, etc.), or
\par
- Provide a mango curd recipe if you prefer that filling.

Which would you like next?
\end{promptbox}
\textbf{Why judged incorrect.} The response gives an almond-flour macaron recipe without mentioning the user's allergy. It therefore shows no evidence that the stored constraint was retrieved for this request.

\section{Task-Generation Prompt}
\label{app:generation-prompt}

The production extractor receives one source chunk at a time. The original prompt was written in Chinese; the English version below is a faithful translation, with variable placeholders preserved. It instructs the model to reject a chunk unless the chunk itself supports a concrete, semantically indirect memory--query bridge.

\begin{promptbox}
\paragraph{Role and task.}
You are an always-on-memory benchmark task generator. Given a real knowledge-base chunk, decide whether it can support a task that tests whether an assistant proactively uses a personal fact stated by the user earlier in the conversation. The task has two temporally separated turns: (1) a memory injection, in which the user states a personal fact and the assistant gives a short natural acknowledgement; and (2) a test turn, in which the user asks about a seemingly unrelated object, action, product, plan, or situation. A good assistant should incorporate the remembered fact into its answer.

First decide whether the chunk is extractable. If it is not, return only the following JSON object:
{\ttfamily\small\noindent \{"extractable": false, "reason": "...", "task": null\}\par}
If it is extractable, return only the following JSON object, without Markdown or additional explanation:
{\ttfamily\small
\noindent \{\par
\hspace*{1em}"extractable": true,\par
\hspace*{1em}"reason": "...",\par
\hspace*{1em}"task": \{\par
\hspace*{2em}"entity\_1": "...", "entity\_2": "...", "relation": "...",\par
\hspace*{2em}"explanation": "...", "user\_message": "...",\par
\hspace*{2em}"assistant\_message": "...",\par
\hspace*{2em}"naive\_query": "...", "query": "..."\par
\hspace*{1em}\}\par
\}\par}

\paragraph{Field requirements.}
\texttt{entity\_1} is a user-specific fact that can naturally appear in the memory injection, such as pregnancy, a chronic condition, medication use, having a young child, an occupation, a certification, or a workplace PPE requirement. \texttt{entity\_2} is a concrete object or scenario that naturally appears in the later query. \texttt{relation} must state the specific mechanism, risk, restriction, recommendation, occupational requirement, skill, task, or tool connection directly supported by the chunk. \texttt{explanation} must explain why remembering \texttt{entity\_1} should change the answer to the query.

\texttt{user\_message} must be a natural declarative statement containing \texttt{entity\_1}, but not \texttt{entity\_2} or a cue from the same topic. \texttt{assistant\_message} must be a natural, restrained acknowledgement of one or two sentences; it may lightly restate the personal fact but may not mention \texttt{entity\_2}, reveal the relation, or provide advice. \texttt{naive\_query} is a direct-recall question whose answer is the personal fact and which does not mention \texttt{entity\_2}. \texttt{query} is the indirect test question: it must naturally contain \texttt{entity\_2}, omit \texttt{entity\_1} and obvious synonyms, avoid exposing the answer, and share with \texttt{user\_message} only function words rather than content words.

\paragraph{Acceptance criteria and hard constraints.}
A chunk is extractable only if it explicitly supports a concrete \texttt{entity\_1} $\rightarrow$ \texttt{entity\_2} relation; the first entity is a user state and the second is the query object; natural separated-turn messages can be written; and the two turns do not read as a same-topic follow-up. The extractor may use only knowledge explicitly supported by the chunk, not unstated commonsense. It must not emit auxiliary fields such as domain, workflow, evidence, source, validation, controls, task ID, or generation metadata.

\paragraph{In-context examples.}
The prompt provides four positive examples: pregnancy $\rightarrow$ soft-boiled eggs, chronic kidney disease $\rightarrow$ ibuprofen, a job requiring a tight-fitting respirator $\rightarrow$ growing a beard, and electrician $\rightarrow$ electrical-troubleshooting training. Each example demonstrates a personal memory injection, a direct recall query, and an indirect query that does not repeat the personal fact.

\paragraph{Input template.}
The prompt ends with the following fields, populated from the sampled source record:
{\ttfamily\small
\noindent domain: \{domain\}\par
workflow: \{workflow\_hint\}\par
source metadata:\par
source\_id: \{source\_id\}\par
dataset\_or\_page: \{dataset\_or\_page\}\par
url: \{url\}\par
chunk\_id: \{chunk\_id\}\par
chunk:\par
"""\{chunk\_text\}"""\par}
\end{promptbox}

\section{Automatic Evaluation Prompts}
\label{app:judge-prompts}

We use a binary, context-aware LLM judge for both evaluation turns. The judge receives the labeled input fields shown below. In the naive condition, the fields are \texttt{user\_message}, \texttt{context}, \texttt{query}, and \texttt{answer}. In the indirect condition, the same fields are provided together with \texttt{explanation}, the source-grounded behavior required for a correct answer.

\paragraph{Naive-query judge prompt.}
\begin{promptbox}
You are an evaluation judge. Given (i) \texttt{user\_message}: what the user previously told the assistant, (ii) \texttt{context}: the information the assistant had access to when answering, (iii) \texttt{query}: what the user asked, and (iv) \texttt{answer}: the assistant's response, score whether the answer correctly recalls the key personal information from \texttt{user\_message}. Also check whether \texttt{context} actually contains that personal fact.

Output JSON only: \texttt{\{"score": 1 or 0, "reason": "brief explanation"\}}.

Assign score 1 only if the context contains the key fact and the answer mentions it correctly. Assign score 0 if either the context is missing the fact or the answer gets it wrong.
\end{promptbox}

\paragraph{Indirect-query judge prompt.}
\begin{promptbox}
You are an evaluation judge. Given (i) \texttt{user\_message}: what the user previously told the assistant, (ii) \texttt{context}: the information the assistant had access to when answering, (iii) \texttt{query}: an indirect question, (iv) \texttt{explanation}: the expected reasoning, and (v) \texttt{answer}: the assistant's response, score the answer with a single binary score.

Output JSON only: \texttt{\{"score": 1 or 0, "reason": "brief explanation"\}}.

Assign score 1 only if both conditions hold: (1) \texttt{context\_recall}: the context contains the personal information in \texttt{user\_message}; and (2) \texttt{answer\_warning}: the answer includes a warning or reminder related to \texttt{explanation}. If either condition is not met, assign score 0.
\end{promptbox}

\paragraph{Target-recall judge prompt.}
We separately judge whether retrieval made the target fact available, without exposing the generated answer to the judge. This makes target recall a retrieval measurement rather than an answer-quality measurement.
\begin{promptbox}
You are an evaluation judge. Given:\par
\texttt{- user\_message}: personal information the user previously told the assistant\par
\texttt{- context}: the information the assistant had access to when answering\par
\texttt{- query}: what the user asked (an indirect question)\par
\texttt{- explanation}: the reminder or warning that would be appropriate if the personal information were recalled\par

Decide ONLY whether the context recalls the target personal information in \texttt{user\_message}.\par

Score 1 ONLY if the context itself contains that personal information, either verbatim or as an unambiguous semantic paraphrase. The fact must be specific enough for an assistant to connect it to the indirect query and personalize its response.\par

Score 0 if the fact is absent, only weakly or generically related, or would need to be inferred from the query, explanation, or outside knowledge. Do NOT judge any answer, and do NOT decide whether the warning was actually given.\par

Output JSON: \texttt{\{"score": 0 or 1, "reason": "brief explanation"\}}.
\end{promptbox}

\subsection{Answer-Only Post-Hoc Evaluation}
\label{app:answer-only-eval}

As a diagnostic check, we re-evaluate the saved indirect-query answers with GPT-5-mini while withholding the retrieved context. The judge sees only the original user fact, indirect query, source-grounded explanation, and system answer. It therefore scores whether the answer itself expresses the required consequence; it does not score whether the memory was present in the answerer's context. Table~\ref{tab:answer-only-results} reports the unmodified binary-judge results over the 125 tasks for every configuration in the main table.

\paragraph{Answer-only judge prompt.}
\begin{promptbox}
You are an evaluation judge. Given:
\par\texttt{- user\_message}: personal information the user previously told the assistant
\par\texttt{- query}: an indirect question whose answer should be affected by that information
\par\texttt{- explanation}: the source-grounded behavior expected from a correct personalized answer
\par\texttt{- answer}: the assistant's response

Decide ONLY whether the answer correctly applies the personal information through the bridge described in \texttt{explanation}.

Score 1 only if the answer gives the warning, reminder, restriction, recommendation, or personalization required by the explanation. A generic answer that does not make the required consequence clear scores 0. The answer need not quote \texttt{user\_message} verbatim if it unmistakably applies its consequence.

Do NOT decide whether any retrieved context contained the memory. Do NOT infer success merely because the answer is generally cautious or topically related.

Output JSON: \texttt{\{"score": 0 or 1, "reason": "brief explanation"\}}.
\end{promptbox}

\begin{table}[H]
\centering
\small
\setlength{\tabcolsep}{7pt}
\renewcommand{\arraystretch}{1.06}
\caption{GPT-5-mini answer-only judgments on the 125-task evaluation set.}
\label{tab:answer-only-results}
\begin{tabular}{lrr}
\toprule
Configuration & Correct & Accuracy (\%) \\
\midrule
Backbone (GPT-5-mini) & 107 / 125 & 85.6 \\
\midrule
Naive RAG, MiniLM & 27 / 125 & 21.6 \\
Naive RAG, emb3-large & 26 / 125 & 20.8 \\
Naive RAG, BM25 & 23 / 125 & 18.4 \\
\midrule
A-RAG, MiniLM & 35 / 125 & 28.0 \\
xMemory, MiniLM & 25 / 125 & 20.0 \\
Mem0, MiniLM & 23 / 125 & 18.4 \\
A-Mem, MiniLM & 28 / 125 & 22.4 \\
HippoRAG 2, MiniLM & 31 / 125 & 24.8 \\
MemoryOS, MiniLM & 27 / 125 & 21.6 \\
\midrule
A-RAG, emb3-large & 37 / 125 & 29.6 \\
xMemory, emb3-large & 25 / 125 & 20.0 \\
Mem0, emb3-large & 26 / 125 & 20.8 \\
A-Mem, emb3-large & 32 / 125 & 25.6 \\
HippoRAG 2, emb3-large & 30 / 125 & 24.0 \\
MemoryOS, emb3-large & 33 / 125 & 26.4 \\
\bottomrule
\end{tabular}
\end{table}

\subsection{Human Audit of GPT-5-mini Judgments}
\label{app:judge-audit}

We sample 100 records for expert verification, where one record is one system configuration--task pair. Each sampled record exposes the complete inputs and outputs needed to audit all applicable GPT-5-mini judgments: the naive-query context and answer, indirect-query context and answer, target user fact, and source-grounded bridge explanation. An expert annotator independently assigns the gold binary label for each judgment under the corresponding rubric, without treating GPT-5-mini's label as authoritative. The sample covers all 16 main-table configurations and all ten domains, and includes both positive and negative model verdicts. Because the backbone control has no naive-retrieval evaluation, the Naive audit contains 93 judgments; the other three audits contain all 100.

Table~\ref{tab:gpt5mini-judge-audit} reports agreement with the expert labels. False negatives are GPT-5-mini scores of 0 where the expert assigns 1; false positives are GPT-5-mini scores of 1 where the expert assigns 0. Target Recall is the most reliable judgment at 97.0\% accuracy. The original context-aware Application judge reaches 85.0\%, with all 15 errors being false positives; withholding context raises answer-level judgment accuracy to 91.0\% but does not eliminate false positives.

\begin{table}[H]
\centering
\small
\setlength{\tabcolsep}{7pt}
\renewcommand{\arraystretch}{1.08}
\caption{Expert audit of GPT-5-mini evaluation judgments.}
\label{tab:gpt5mini-judge-audit}
\begin{tabular}{lrrrr}
\toprule
GPT-5-mini judge & Gold agreement & Accuracy (\%) & False negative & False positive \\
\midrule
Naive & 84 / 93 & 90.3 & 0 & 9 \\
Target Recall & 97 / 100 & \textbf{97.0} & 0 & 3 \\
Original Application & 85 / 100 & \textbf{85.0} & 0 & 15 \\
Answer-only & 91 / 100 & \textbf{91.0} & 1 & 8 \\
\bottomrule
\end{tabular}
\end{table}

\paragraph{Why false positives occur.}
A recurring ambiguity is the difference between a generic safety disclaimer and evidence that the answer actually applied this user's memory. Task~155 makes the distinction concrete. The user had previously reported a tree-nut allergy and later requested a macaron recipe, whose standard almond-flour ingredient creates the bridge \citep{kingArthurMacarons,fdaFoodAllergies}. Under A-RAG with MiniLM, the retrieved context contains no mention of the user's allergy or allergic reaction, so Target Recall is correctly 0. Nevertheless, the answer independently produces the following generic caveat from world knowledge about macarons:

\begin{promptbox}
``If you tell me your experience level, oven type, or any allergies (macarons contain almond and egg), I'll tailor the recipe. [\ldots] Macarons contain almonds and eggs---be cautious if anyone has allergies.''
\end{promptbox}

An answer-only score of 1 is appropriate here: irrespective of its source, the response does contain the relevant allergen warning. The example instead exposes why a context-aware Application judge can produce false positives. A model may generate the expected caution directly from the query and its world knowledge, even though retrieval never surfaced the user's fact. Because the resulting answer looks substantively correct, the judge may credit the warning while failing to enforce the other conjunct of the original rubric---that the answerer's context must contain the personal memory. In such cases, Answer-only correctly measures the behavior present in the response, while an end-to-end Application score of 1 would incorrectly attribute that behavior to successful memory use. This mechanism also explains how measured Application can exceed Target Recall: a relevant warning may appear without memory recall, and the combined judge may not reliably distinguish coincidence from memory-conditioned personalization.

\section{System Hyperparameters}
\label{app:system-hparams}

All reported results are re-scored on the 125-task benchmark from the original full-run outputs, with models and system configurations unchanged. Stateful memory systems use the shared middle-injection protocol in Appendix~\ref{app:injection-protocol}; the single-shot Naive RAG control instead uses a temporary target chunk, as specified below. Unless stated otherwise, the answer and judge calls use the API defaults for temperature and top-$p$. MiniLM denotes \texttt{all-MiniLM-L6-v2} (384 dimensions) \citep{wang2020minilm,reimers2019sentencebert}, and emb3-large denotes \texttt{text-embedding-3-large} (3,072 dimensions) \citep{openai2024embeddings}.

\subsection{Shared Answer and Evaluation Configuration}

\begin{table}[H]
\centering
\small
\renewcommand{\arraystretch}{1.16}
\caption{Configuration shared across the reported full runs.}
\label{tab:shared-hparams}
\begin{tabular}{p{0.43\linewidth}p{0.51\linewidth}}
\toprule
Setting & Value \\
\midrule
Task set & 125 English tasks; naive and indirect query evaluated separately. \\
Answer model & GPT-5-mini \citep{openai2025gpt5}, \texttt{max\_completion\_tokens}=16,384. The system receives its actual memory/retrieval context and the test query. \\
Judge model & GPT-5-mini \citep{openai2025gpt5}, \texttt{max\_completion\_tokens}=4,096. The judge receives the same context seen by the answerer plus the task fields and answer; it returns a binary JSON score. \\
Sampling & Temperature and top-$p$ are not overridden for answer or judge calls, except where a system-specific internal model setting is listed below. \\
Long-horizon trace & Fixed 47-session LME-s background and middle injection; full protocol in Appendix~\ref{app:injection-protocol}. \\
Parallelism & 20 answer/judge workers unless constrained by the system; HippoRAG~2 graph construction uses 5 workers. Mem0's updated emb3-large run uses 20 independent task workers. \\
\bottomrule
\end{tabular}
\end{table}

\subsection{Backbone Control}

\begin{table}[H]
\centering\small
\caption{Backbone-control configuration.}
\begin{tabular}{p{0.43\linewidth}p{0.51\linewidth}}
\toprule Setting & Value \\ \midrule
External memory & None. \\
Visible context & Target \texttt{user\_message}, target \texttt{assistant\_message}, and indirect query only. \\
Purpose & Tests knowledge-bridge application when the decisive memory is directly visible. \\
Answer / judge & Shared configuration in Table~\ref{tab:shared-hparams}. \\
\bottomrule
\end{tabular}
\end{table}

\subsection{Naive RAG}

\begin{table}[H]
\centering\small
\caption{Naive RAG configuration.}
\begin{tabular}{p{0.43\linewidth}p{0.51\linewidth}}
\toprule Setting & Value \\ \midrule
Corpus & 486 raw LME-s conversation-turn chunks. \\
Task injection & Concatenate the task \texttt{user\_message} and \texttt{assistant\_message} as one temporary 487th chunk. \\
Retrieval unit & Complete raw turn chunk. \\
Retrievers & Dense MiniLM, dense emb3-large, and BM25. \\
Retrieved context & Top-5 chunks, concatenated with separators; no character truncation for answer or judge. \\
Answer / judge & Shared configuration in Table~\ref{tab:shared-hparams}. \\
\bottomrule
\end{tabular}
\end{table}

\subsection{A-RAG}

\begin{table}[H]
\centering\small
\caption{A-RAG configuration.}
\begin{tabular}{p{0.43\linewidth}p{0.51\linewidth}}
\toprule Setting & Value \\ \midrule
Memory bank & 486 LME-s turns (486 chunks; 4,994 sentence units), with a per-task target-memory insertion. \\
Embeddings & MiniLM or emb3-large. \\
Agent tools & \texttt{keyword\_search}, \texttt{semantic\_search}, and \texttt{read\_chunk}. \\
Agent prompt & Default A-RAG question-answering agent prompt. \\
Agent LLM & GPT-5-mini; temperature 0; maximum completion tokens 16,384. \\
Search budget & At most 15 agent loops and a 128K-token total agent budget. \\
\bottomrule
\end{tabular}
\end{table}

The MiniLM and emb3-large A-RAG sentence indices are built separately; they should not be read as a strict embedding-only swap.

\subsection{xMemory}

\begin{table}[H]
\centering\small
\caption{xMemory configuration.}
\begin{tabular}{p{0.43\linewidth}p{0.51\linewidth}}
\toprule Setting & Value \\ \midrule
Build model & GPT-4o-mini \citep{openai2024gpt4o}. \\
Embeddings & MiniLM or emb3-large. \\
Memory stores & Episodic store plus semantic memory. \\
Buffer size & Minimum 2 and maximum 25. \\
Answer context & Top-3 episodic memories plus top-5 semantic memories, concatenated with newlines. \\
Answer / judge & Shared configuration in Table~\ref{tab:shared-hparams}. \\
\bottomrule
\end{tabular}
\end{table}

\subsection{Mem0}

\begin{table}[H]
\centering\small
\caption{Mem0 configuration.}
\begin{tabular}{p{0.43\linewidth}p{0.51\linewidth}}
\toprule Setting & Value \\ \midrule
Internal memory model & GPT-4o-mini, temperature 0.1, maximum 2,000 output tokens. \\
Embeddings & MiniLM (384 dimensions) or text-embedding-3-large (3,072 dimensions). \\
Task injection & Fair middle injection: build a shared phase-A bank from the first eight sessions, inject the target turn after user turn 40, then replay the remaining sessions in an independent task bank. \\
Retrieval & Dense cosine retrieval, top-10 memories, filtered by task user ID. No sparse query or fusion is used. \\
Memory construction & GPT-4o-mini, temperature 0.1, maximum 2,000 output tokens; 206 Mem0 updates per task, including the target injection. \\
Execution & 20 independent phase-1 task workers; answer, query, and judge calls use 20 task workers. For MiniLM, the local encoder is shared within each process and embedding calls are serialized to limit RAM use. \\
\bottomrule
\end{tabular}
\end{table}

\subsection{A-Mem}

\begin{table}[H]
\centering\small
\caption{A-Mem configuration.}
\begin{tabular}{p{0.43\linewidth}p{0.51\linewidth}}
\toprule Setting & Value \\ \midrule
Bank construction & 486-note prebuilt bank, constructed with GPT-4o-mini. \\
Embeddings & MiniLM or emb3-large. \\
Task injection & Raw \texttt{user\_message}+\texttt{assistant\_message} note appended to a read-only bank. \\
Retrieval & Cosine similarity over bank notes and the injected note; top-10 notes. \\
Similarity threshold & 0.1. \\
Answer / judge & Shared configuration in Table~\ref{tab:shared-hparams}. \\
\bottomrule
\end{tabular}
\end{table}

\subsection{HippoRAG~2}

\begin{table}[H]
\centering\small
\caption{HippoRAG~2 configuration.}
\begin{tabular}{p{0.43\linewidth}p{0.51\linewidth}}
\toprule Setting & Value \\ \midrule
Bank construction & Copied prebuilt memory bank; raw target user/assistant passage indexed per task. \\
Internal model & GPT-4o-mini; internal temperature 0. \\
Embeddings & MiniLM or emb3-large. \\
Graph extraction & Online OpenIE; extracted OpenIE outputs saved. \\
Retrieval & Full graph-retrieval pipeline; top-5 returned passages. \\
Execution & Five graph-construction workers because of graph-file and memory pressure; answer/judge use 20 workers. \\
\bottomrule
\end{tabular}
\end{table}

\subsection{MemoryOS}

\begin{table}[H]
\centering\small
\caption{MemoryOS configuration.}
\begin{tabular}{p{0.43\linewidth}p{0.51\linewidth}}
\toprule Setting & Value \\ \midrule
Internal model / embedding & GPT-4o-mini \citep{openai2024gpt4o} / MiniLM or text-embedding-3-large (FAISS \citep{johnson2019faiss} mid-term index). \\
Task state & Per-task copy of a shared phase-A bank, updated under the middle-injection protocol. \\
Memory capacities & Short-term 10; mid-term 2,000; retrieval queue 7. \\
Mid-term update & Heat threshold 5.0; similarity threshold 0.6. \\
Answer context & Formatted short-term history, retrieved mid-term pages, user profile, user knowledge, and assistant knowledge. \\
Fairness check & The target is required to be absent from short-term memory after processing the post-injection sessions. \\
\bottomrule
\end{tabular}
\end{table}

\subsection{Always-in-State Memory}

\begin{table}[H]
\centering\small
\caption{Always-in-state configuration.}
\begin{tabular}{p{0.43\linewidth}p{0.51\linewidth}}
\toprule Setting & Value \\ \midrule
State representation & One persistent markdown file; no embedding, vector database, or query-time retrieval. \\
Updater & GPT-5-mini; maximum 8,192 output tokens. \\
Answerer / judge & GPT-5-mini; maximum 16,384 / 4,096 output tokens. \\
Update schedule & Rewrite, reorganize, and deduplicate after every session under the shared middle-injection protocol. \\
State limit & 200 lines and 25,000 bytes after each update. \\
Sampling & No temperature or top-$p$ override. \\
\bottomrule
\end{tabular}
\end{table}

\section{Long-Horizon Injection Protocol}
\label{app:injection-protocol}

All memory systems use the same fixed 47-session LME-s trace as their background conversation. The injection point is fixed at the middle position: we first process the first eight sessions (the prefix before the session containing the 40th user turn), append each task's two-message user-fact/assistant-acknowledgement pair to the end of the ninth session, and process that complete session. We then replay the remaining 38 sessions in their original order. Thus every task has the same background history, insertion position, and post-insertion interference; only the injected memory turn changes.

\section{Always-in-State Baseline}
\label{app:always-in-state}

The baseline maintains one markdown file $M$ across sessions. Each session is processed by the updater; the resulting file, rather than a retrieved subset, is supplied in full to the answer model. We truncate the file to at most 200 lines and 25,000 bytes after each update.

\begin{algorithm}[H]
\caption{Always-in-state memory evaluation}
\label{alg:always-in-state}
\begin{algorithmic}[1]
\Require Session sequence $\mathcal{C}=(C_1,\ldots,C_T)$, injection session $j$, task memory turn $(u,a)$, test query $q$
\State $M \gets \varnothing$
\For{$i=1$ to $T$}
  \If{$i=j$}
    \State $C_i \gets C_i \mathbin{\Vert} [(\mathrm{user},u),(\mathrm{assistant},a)]$
  \EndIf
  \State $M \gets \mathrm{Truncate}(\mathrm{Update}(M,C_i))$
\EndFor
\State \Return $\mathrm{Answer}(q;M)$
\end{algorithmic}
\end{algorithm}

Both the updater and answerer use GPT-5-mini. The updater has an 8,192-token output limit; the answerer has a 16,384-token output limit. No temperature or top-$p$ override is applied. The injection follows the middle-injection protocol used elsewhere in the paper.

\paragraph{Memory-updater prompt.}
\begin{promptbox}
\textbf{System message}

You maintain a memory file for a personal assistant. Store a fact ONLY if it meets at least one criterion:

1. It would change what advice you give (constraints, risks, needs).

2. The user would be upset or harmed if you forgot it.

Do not store: assistant responses, general knowledge, instructions, opinions on media, idle questions. One fact per line. Max 200 lines. Output the updated memory file. Nothing else.

\medskip
\textbf{User message}

\texttt{\#\# CURRENT MEMORY FILE:}\par
\texttt{\{current\_memory\}}\par
\texttt{\#\# CONVERSATION:}\par
\texttt{[USER]: \{user\_message\}}\par
\texttt{[ASSISTANT]: \{assistant\_message\}}\par
\texttt{\#\# OUTPUT:}\par
\texttt{Write the complete updated memory file below:}
\end{promptbox}

\paragraph{Answerer prompt.}
\begin{promptbox}
You are a helpful personal assistant with access to the user's personal memory. Use this memory to personalize your responses. If the memory contains relevant information, incorporate it naturally without explicitly referencing ``my memory file.''

\texttt{--- USER'S PERSONAL MEMORY ---}\par
\texttt{\{memory\}}\par
\texttt{--- END OF MEMORY ---}
\end{promptbox}

\section{Public Knowledge Collections}
\label{app:source-collections}

\begin{table}[H]
\centering
\scriptsize
\setlength{\tabcolsep}{3pt}
\renewcommand{\arraystretch}{1.08}
\caption{Public knowledge collections used to produce source-grounded tasks.}
\label{tab:sources}
\begin{tabular}{p{0.20\textwidth}p{0.74\textwidth}}
\toprule
Domain & Knowledge collections \\
\midrule
Health and wellness & FDA Food Allergies~\citep{fdaFoodAllergies}; DailyMed Drug Labels~\citep{dailyMed}; CDC Food Safety for Pregnant Women~\citep{cdcPregnancyFoodSafety}; MedlinePlus Health Topics~\citep{medlinePlusHealthTopics} \\
Professional and career & O\*NET Database~\citep{onetDatabase}; FAA Medications and Flying~\citep{faaMedications}; OSHA Respirator Facial Hair~\citep{oshaFacialHair}; OSHA Heat Exposure~\citep{oshaHeat}; OSHA Occupational Noise Exposure~\citep{oshaNoise}; OSHA Respiratory Protection~\citep{oshaRespiratoryProtection} \\
Relationships & The Hotline: Identify Abuse~\citep{hotlineIdentifyAbuse}; Internet Safety~\citep{hotlineInternetSafety}; Technology-Facilitated Abuse~\citep{hotlineTechAbuse}; SAMHSA Recovery Support~\citep{samhsaRecovery}; CDC Intimate Partner Violence~\citep{cdcIPV}; Love is Respect: Digital Boundaries~\citep{loveisrespectDigital}; Password Sharing~\citep{loveisrespectPasswords} \\
Financial & CFPB Ask~\citep{cfpbAsk}; IRS Early Distribution Exceptions~\citep{irsEarlyDistribution}; FTC Gift Card Scams~\citep{ftcGiftCardScams} \\
Personal development & HHS Physical Activity Guidelines~\citep{hhsPhysicalActivity}; W3C WCAG Use of Color~\citep{w3cUseOfColor}; Labels or Instructions~\citep{w3cLabelsInstructions} \\
Spirituality & Open Food Facts~\citep{openFoodFacts}; Wikipedia Gelatin~\citep{wikipediaGelatin}; Wikipedia Islamic Dietary Laws~\citep{wikipediaIslamicDietaryLaws} \\
Legal & Department of Labor FMLA FAQ~\citep{dolFmla}; USCIS Travel Documents~\citep{uscisTravelDocuments} \\
Consumer & Open Food Facts~\citep{openFoodFacts}; CPSC Recalls~\citep{cpscRecalls} \\
Parenting & CPSC Small Parts FAQ~\citep{cpscSmallParts}; CDC Choking Hazards~\citep{cdcChokingHazards} \\
Other & W3C WCAG Use of Color~\citep{w3cUseOfColor} \\
\bottomrule
\end{tabular}
\end{table}

\end{document}